\RequirePackage{fix-cm}
\documentclass[twocolumn]{svjour3}          
\smartqed  
\usepackage{graphicx}
\usepackage{epsfig}
\usepackage{amsmath}
\usepackage{amssymb}
\usepackage{natbib}
\usepackage[colorlinks,citecolor=blue]{hyperref}

\usepackage{amsmath}
\usepackage{amssymb,subfigure,cases}
\usepackage{setspace}
\usepackage{color,hyperref}
\usepackage{algpseudocode}
\usepackage{booktabs} 

\usepackage{times}
\usepackage{amssymb}
\usepackage{multirow}
\usepackage{subfigure}
\usepackage{pifont}
\usepackage{xcolor}
\usepackage{array}
\usepackage{url}
\usepackage{soul}

\newcommand{\xmark}{\ding{55}}
\newcommand{\rmark}{\ding{52}}

\newcolumntype{I}{!{\vrule width 1.2pt}}
\newlength\savedwidth
\newcommand\whline{\noalign{\global\savedwidth\arrayrulewidth
		\global\arrayrulewidth 1.25pt}%
	\hline
	\noalign{\global\arrayrulewidth\savedwidth}}
\newcommand{\tabincell}[2]{\begin{tabular}{@{}#1@{}}#2\end{tabular}}

\begin{document}
	
\setul{}{1.2pt}

\title{Pixel-wise Crowd Understanding via Synthetic Data 
}


\author{Qi Wang        \and
	Junyu Gao \and
	Wei Lin  \and 
	Yuan Yuan*
}


\institute{
	* Y. Yuan is the corresponding author. \\
	Q. Wang, J. Gao, W. Lin and Y. Yuan are with the School of Computer Science and with the Center for Optical Imagery Analysis and Learning (OPTIMAL), Northwestern Polytechnical University, Xi'an 710072, Shaanxi, China. E-mails: crabwq@gmail.com, gjy3035@gmail.com, elonlin24@gmail.com, y.yuan1.ieee@gmail.com.
}

\date{Received: date / Accepted: date}

\maketitle

\begin{abstract}

Crowd analysis via computer vision techniques is an important topic in the field of video surveillance, which has wide-spread applications including crowd monitoring, public safety, space design and so on. Pixel-wise crowd understanding is the most fundamental task in crowd analysis because of its finer results for video sequences or still images than other analysis tasks. Unfortunately, pixel-level understanding needs a large amount of labeled training data. Annotating them is an expensive work, which causes that current crowd datasets are small. As a result, most algorithms suffer from over-fitting to varying degrees. In this paper, take crowd counting and segmentation as examples from the pixel-wise crowd understanding, we attempt to remedy these problems from two aspects, namely data and methodology. Firstly, we develop a free data collector and labeler to generate synthetic and labeled crowd scenes in a computer game, Grand Theft Auto V. Then we use it to construct a large-scale, diverse synthetic crowd dataset, which is named as ``GCC Dataset''. Secondly, we propose two simple methods to improve the performance of crowd understanding via exploiting the synthetic data. To be specific, 1) supervised crowd understanding: pre-train a crowd analysis model on the synthetic data, then fine-tune it using the real data and labels, which makes the model perform better on the real world; 2) crowd understanding via domain adaptation: translate the synthetic data to photo-realistic images, then train the model on translated data and labels. As a result, the trained model works well in real crowd scenes. 

Extensive experiments verify that the supervision algorithm outperforms the state-of-the-art performance on four real datasets: UCF\_CC\_50, UCF-QNRF, and Shanghai Tech Part A/B Dataset. The above results show the effectiveness, values of synthetic GCC for the pixel-wise crowd understanding. The tools of collecting/labeling data, the proposed synthetic dataset and the source code for counting models are available at \url{https://gjy3035.github.io/GCC-CL/}.
	



\end{abstract}

\section{Introduction}
\label{intro}

\begin{figure*}
	\centering
	\includegraphics[width=0.85\textwidth]{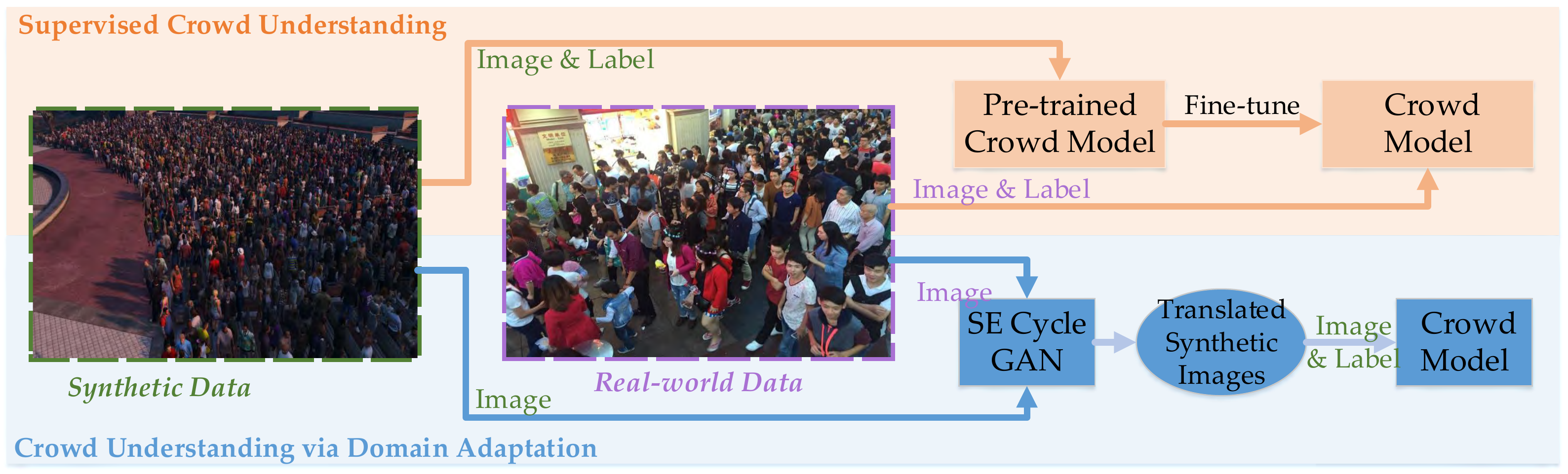}
	\caption{Two strategies of using the synthetic GCC dataset for pixel-wise crowd understanding: supervised learning and domain adaptation. The former firstly trains a pre-trained model on GCC and then fine-tune the model on real-world data. This strategy is able to significantly improve performance of the traditional supervise methods. The latter firstly adopts a CycleGAN-based method to translate the GCC data to photo-realistic scenes, then the trained crowd model uses the translated data and labels. The entire domain adaptation process does not need any label of real data.}\label{Fig-intro}
\end{figure*}

Recently, crowd analysis has been a hot topic in the field of computer vision. It has great potential (including visual surveillance, crowd management, public space design and so on) in the real-world crowd scenes: railway station, shopping mall, stadium, airport terminals, theaters, public buildings, etc. \citep{chan2009analysis,li2014crowded}. Some typical crowd analysis tasks includes crowd counting/density estimation \citep{chan2008privacy,chan2009bayesian,idrees2013multi,wan2019residual,yan2019perspective}, crowd segmentation \citep{dong2007fast,kang2014fully}, crowd anomaly detection \citep{mahadevan2010anomaly,li2013anomaly,yuan2014online}, human behavior recognition \citep{mehran2009abnormal,popoola2012video}, person group detection \citep{li2017multiview,wang2018detecting}, pedestrian tracking \citep{zuo2018learning,li2018visual}. 
To be specific, crowd counting and segmentation are two essential tasks in crowd analysis, which are pixel-wise regression and classification task. In this paper, the both are together treated as \textbf{Pixel-wise Crowd Understanding}. The former aims to estimate the crowd density for each region in scenes and produces the number of people. The latter can predict the crowd region at the pixel level. Accurate density estimation and crowd localization provide some important attention information and fine pixel-wise results for other high-level analysis tasks. For example, semantic person group detection needs preliminary crowd segmentation results. Thus, pixel-wise crowd understanding is the most fundamental task in the field of crowd analysis.

With the development of deep learning, many methods \citep{onororubio2016towards,walach2016learning,xiong2017spatiotemporal,sam2018divide,shen2018crowd,shi2018crowd} based on CNN attain a remarkable performance in the pixel-wise understanding task on the current research datasets. However, most algorithms only focus on how to learn effective and discriminative features (such as texture patterns, local structural features, global contextual information, multi-scale features and so on) to improve models' performance in the specific dataset but ignore the results in the wild. In fact, there is a large performance degradation during deploying them to the wild or other real-world scenes. The essential reason is that the model over-fits the scarce data. 

For the scarce data problem, extending data seems like a very straightforward and effective solution. Unfortunately, annotating pixel-wise data is an expensive task: accurately labeling a scene contains 1,000 people will take more than 40 minutes. Thus, current crowd datasets \citep{chan2008privacy, chen2012feature, zhang2016data, zhang2016single,Qi2017Deep,idrees2013multi,idrees2018composition} are small data volume so that they can not perfectly satisfy the needs of the mainstream CNN-based methods. Take the congested crowd dataset as an example, NWPU-Crowd contains only 5,106 images. In addition, there are disadvantages in the existing crowd dataset. Firstly, for extremely congested scenes, the labels of head locations are not very accurate and some heads are missed, such as some samples in UCF\_CC\_50  and Shanghai Tech Part A  (``SHT A'' for short). Secondly, existing datasets lack scenes such as night, variant illumination, and a large-range number of people, which are very common in real life. 

Therefore, we first try to reduce the above problems from the data point of view. Our goal is to construct a large-scale, diverse, low-cost crowd dataset. To this end, with the help of a game engine, we develop a data collector and labeler in a popular computer game (Grand Theft Auto V, GTA V for short, an electronic game developed by Rockstar Games), which can generate synthetic crowd scenes and automatically annotate them. By the proposed collector and labeler, we successfully construct a synthetic crowd dataset, which is named as ``GTA V Crowd Counting'' (``GCC'' for short) dataset. Compared with the existing real-world datasets. GCC dataset has four advantages: 1) free collection and annotation; 2) larger data volume and higher image resolution; 3) more diversified scenes, 4) more accurate annotations and label types (head dot and crowd mask). The detailed statistics are reported in Table \ref{Table-compare}.

After generating the large-scale dataset, then we attempt to exploit it to improve the performance in the wild from the methodological perspective. Here, we have two ways to realize our goal. To be specific, we firstly exploit the synthetic data to pre-train a crowd model, a Fully Convolutional Network, then fine-tune the model using real-world data. This strategy can effectively prompt performance in the real world. Traditional methods (training from scratch \citep{zhang2016single,ranjan2018iterative,cao2018scale} or image classification models \citep{babu2018divide,shi2018crowd,idrees2018composition}) have some layers with random initialization or a regular distribution, which is not a good scheme. Compared with them, our strategy can provide more complete and better initialization parameters. The entire pipeline is shown in the \mbox{\color{orange}{orange}} box of Figure \mbox{\ref{Fig-intro}}. 

The above supervised learning method still needs real-world data and labels. It is an intractable problem that how to get rid of manual labels. Inspired by un-paired image translation \mbox{\citep{zhu2017unpaired}}, we firstly translate the synthetic images to photo-realistic images. Compared with the original CycleGAN, we propose the Structural Similarity Index (SSIM) loss to maintain the texture features and local patterns in the crowd region during the translation process. Then, we train a crowd model via adversarial learning on the labeled translated domain and the unlabeled real domain, which works well in the wild. The flowchart is demonstrated in the \mbox{\textcolor[rgb]{0.3,0.3,1}{light blue}} box of Fig. \mbox{\ref{Fig-intro}}.

In summary, this paper's contributions are four-fold:
\begin{enumerate}
	\item[1)] We are the first to develop a data collector and labeler for crowd counting, which can automatically construct synthetic crowd scenes in GTA V game and simultaneously annotate them without any labor costs. 
	\item[2)] We create the first large-scale and synthetic crowd counting dataset by using the data collector and labeler, which contains 15,212 images, a total of 7,625,843 people. 	
	\item[3)] We present a pre-trained scheme to facilitate the original method's performance on the real data, which can more effectively reduce the estimation errors compared with random initialization and ImageNet model. Further, through the strategy, our proposed SFCN achieves the state-of-the-art results.
	\item[4)] We are the first to propose a crowd understanding method via domain adaptation, which does not use any label of the real data. By our designed SE CycleGAN, the domain gap between the synthetic and real data can be significantly reduced. Finally, the proposed method outperforms the two baselines that face the same problem. 
\end{enumerate}

\begin{table*}[htbp]	
	
	\centering
	\caption{Statistics of the eight real-world datasets and the synthetic GCC dataset. Specifically, the real-world datasets include UCSD  \citep{chan2008privacy}, Mall  \citep{chen2012feature}, UCF\_CC\_50 \citep{idrees2013multi}, WorldExpo'10 \citep{zhang2016data}, SHT A/B \citep{zhang2016single},  UCF-QNRF \citep{idrees2018composition}, NWPU-Crowd \citep{gao2020nwpu} and JHU-CROWD++ \citep{sindagi2019pushing,sindagi2020jhu-crowd++}.}
	\setlength{\tabcolsep}{2.5mm}{
		\begin{tabular}{c|c|c|c|c|c|c|c|c}
			\whline
			\multirow{2}{*}{Dataset}	&Number &Average &\multicolumn{4}{|c}{Count Statistics} &\multicolumn{2}{|c}{Label Forms}\\
			\cline{4-9} 
			& of Images &Resolution & Total &Min & Ave & Max &Head Dot & Crowd Mask	\\
			\whline
			UCSD    &2,000 &$158 \times 238$  & 49,885 &11 & 25 & 46 &\rmark & \xmark	\\
			\hline
			Mall  &2,000 &$480 \times 640$  & 62,325 &13 & 31 & 53 &\rmark & \xmark	\\
			\hline
			UCF\_CC\_50   &50 &$2101 \times 2888$  & 63,974 &94 & 1,279 & 4,543 &\rmark & \xmark	\\
			\hline
			WorldExpo'10 &3,980 &$576 \times 720$  & 199,923 &1 & 50 & 253 &\rmark & \xmark \\
			\hline
			SHT A   &482 &$589 \times 868 $  & 241,677 &33 & 501 & 3,139 &\rmark & \xmark\\
			\hline
			SHT B   &716 &$768 \times 1024$  & 88,488 &9 & 123 & 578 &\rmark & \xmark \\
			\hline
			UCF-QNRF  &1,525 &$2013 \times 2902$  & 1,251,642 & 49 & 815 & 12,865 &\rmark & \xmark \\
			\hline
			NWPU-Crowd  &5,109 & $2311 \times 3383$& 2,133,238 & 0 & 418 & 20,033 &\rmark & \xmark \\
			\hline
			JHU-CROWD++  &4,372 &$910 \times 1430$  & 1,515,005 & 0 & 346 & 25,791 &\rmark & \xmark \\
			\whline
			\textbf{GCC}  &\textbf{15,212} &$\boldsymbol{1080 \times 1920}$  &\textbf{7,625,843}  &\textbf{0} &\textbf{501} &\textbf{3,995} &\rmark & \rmark \\
			\whline
			
		\end{tabular}\label{Table-compare}
	}
\end{table*}

This paper is an extension of our previous work on crowd counting \citep{wang2019learning} in the IEEE Conference on Computer Vision and Pattern Recognition. Compared with the conference version, this paper has some extensions as follows: 
\begin{enumerate}
	\item[1)] \textbf{Data Generation:} The process of scene synthetic is optimized for more efficient data generation, which reduces the simulation time  by two-thirds. Furthermore, we give a more detailed description of the entire process for data generation, including scene selection, setting and synthesis. 
	\item[2)] \textbf{Dataset:} In addition to the head dot labels, the crowd mask for segmentation is provided. It displays pixel-wise salient regions of crowd, which is also an important  fundamental task of crowd analysis.
	\item[3)] \textbf{Methodology:} For supervised crowd understanding, we add a new network to segment the crowd mask based on SFCN. For Crowd understanding via domain adaptation, we add the adversarial learning to jointly train SE CycleGAN and SFCN to further prompt counting performance in the real world.
	\item[4)] \textbf{Experiments:} More further experiments are conducted to verify the effectiveness (improvement of performance, generalization ability, etc.) of the two proposed ways on the real-world datasets.
\end{enumerate}

The rest of this paper is organized as follows. Section \ref{related} reviews the related work briefly in terms of crowd understanding, crowd dataset and sythetic dataset. Section \ref{gcc-dataset} describes the generation process and key features of GCC dataset. Section \ref{SCC} and \ref{DA} respectively focuses on supervised learning and domain adaptation for pixel-wise crowd understanding. Further, the experimental results and discussion are analyzed in Section \ref{exp}. Finally, this work is summarized in Section \ref{conclusion}.

\section{Related Works}
\label{related}

\subsection{Pixel-wise Crowd Understanding} 

Pixel-wise crowd understanding mainly consists of crowd counting and segmentation task. In the past half-decade, researchers exploit Convolutional Neural Network (CNN) to design the effective crowd model, which attains a significant improvement. Some methods \citep{wang2015deep,fu2015fast} attempt to directly regress the number of people for image patches by modifying the traditional CNN classification models. However, there are more semantic gaps in direct regression than pixel-wise density estimation. Thus, many methods \citep{marsden2016fully,liu2018crowd,kang2014fully} adopt Fully Convolutional Networks (FCN) \citep{long2015fully} to produce the pixel-wise density map or predict the crowd region.

Benefiting from FCN's remarkable performance on pixel-wise task (such as semantic segmentation, saliency detection and on), almost all algorithms use FCN to predict crowd density map. Some methods \citep{zhang2016single,idrees2018composition,cao2018scale,ranjan2018iterative} integrate different feature maps from different layers in FCN to improve the quality of density maps. To be specific, \cite{zhang2016single,cao2018scale} design a multi-column CNN and fuses the feature map from different columns to predict the final density map. \cite{idrees2018composition} compute loss from shallow to deep layers by different loss functions to output fine maps. \cite{Jiang_2019_CVPR} present a combinatorial loss to enforce similarities in different spatial scales between prediction maps and groundtruth.
However, it is hard to train a single regression model for density map estimation, which converges slowly and performs not well. Thus, some methods \citep{sindagi2017cnn,8723079,Lian_2019_CVPR,Zhao_2019_CVPR} exploit multi-task learning to explore the relation of different tasks to improve the counting performance, such as high-level density classification, foreground/crowd segmentation, perspective prediction, crowd depth estimation and so on. In addition to multi-task learning, \cite{sindagi2017generating,li2018csrnet,Liu_2019_CVPR} attempt to encode the large-range contextual information via patch-level classification, dilatation convolution and multiple receptive field sizes for crowd scenes. 

For handling scarce training data, \cite{liu2018leveraging} propose a self-supervised learning method to learn to rank a large amount of unlabeled web data, and \cite{shi2018crowd} present a deep negative correlation learning to reduce the over-fitting. \cite{sam2019almost} present an almost unsupervised dense counting autoencoder, in which 99.9\% of parameters are trained without any labeled data.

\subsection{Crowd Datasets} 

In addition to the above algorithms for crowd understanding, the datasets potentially boost the development of crowd counting. The first crowd counting dataset, UCSD \citep{chan2008privacy}, is released by Chan \emph{et al.} from the University of California San Diego, which records a sparse crowd scene in a pedestrian walkway. In addition to counting labels, UCSD also provides identity information, traveling direction and instantaneous velocity for each person. Mall dataset \citep{chen2012feature} is captured from a surveillance camera by Chen \emph{et al.}, containing over 60,000 pedestrian instances in an indoor shopping mall. 

Considering that UCSD and Mall are collected from a single sparse scene, some researchers build extremely congested and diversified-scene crowd counting dataset. \cite{idrees2013multi} release a highly congested crowd dataset named UCF\_CC\_50, containing 50 images. The average number of people per image is more than 1,200. \cite{zhang2016data} construct a cross-scene crowd dataset, WorldExpo'10. It includes 120 different crowd scenes that are captured from surveillance cameras in Shanghai 2010 WorldExpo. \cite{zhang2016single} present ShanghaiTech Dataset, including the high-quality real-world images. It consists of 2 parts: Part A is collected from a photo-sharing website  \footnote{https://www.flickr.com/} and Part B is captured from the walking streets in Shanghai. For covering the more large-range, large-scale crowd scene and more accurate annotation, \cite{idrees2018composition} propose a highly congested dataset with higher resolution, UCF-QNRF, of which number range if from 49 to 12,865. UCF-QNRF is by far the largest extremely congested crowd counting dataset, containing 1,525 images, in total of 1,251,642 persons. Recently, JHU and NWPU release two large-scale crowd counting datasets, \mbox{\citep{gao2020nwpu,sindagi2019pushing,sindagi2020jhu-crowd++}}, which will promote the further development of the community of crowd counting. 

However, the above datasets suffer some drawbacks mentioned in Section \ref{intro} to some extent. More detailed information about them is shown in Table \ref{Table-compare}.

\begin{figure*}
	\centering
	\includegraphics[width=1\textwidth]{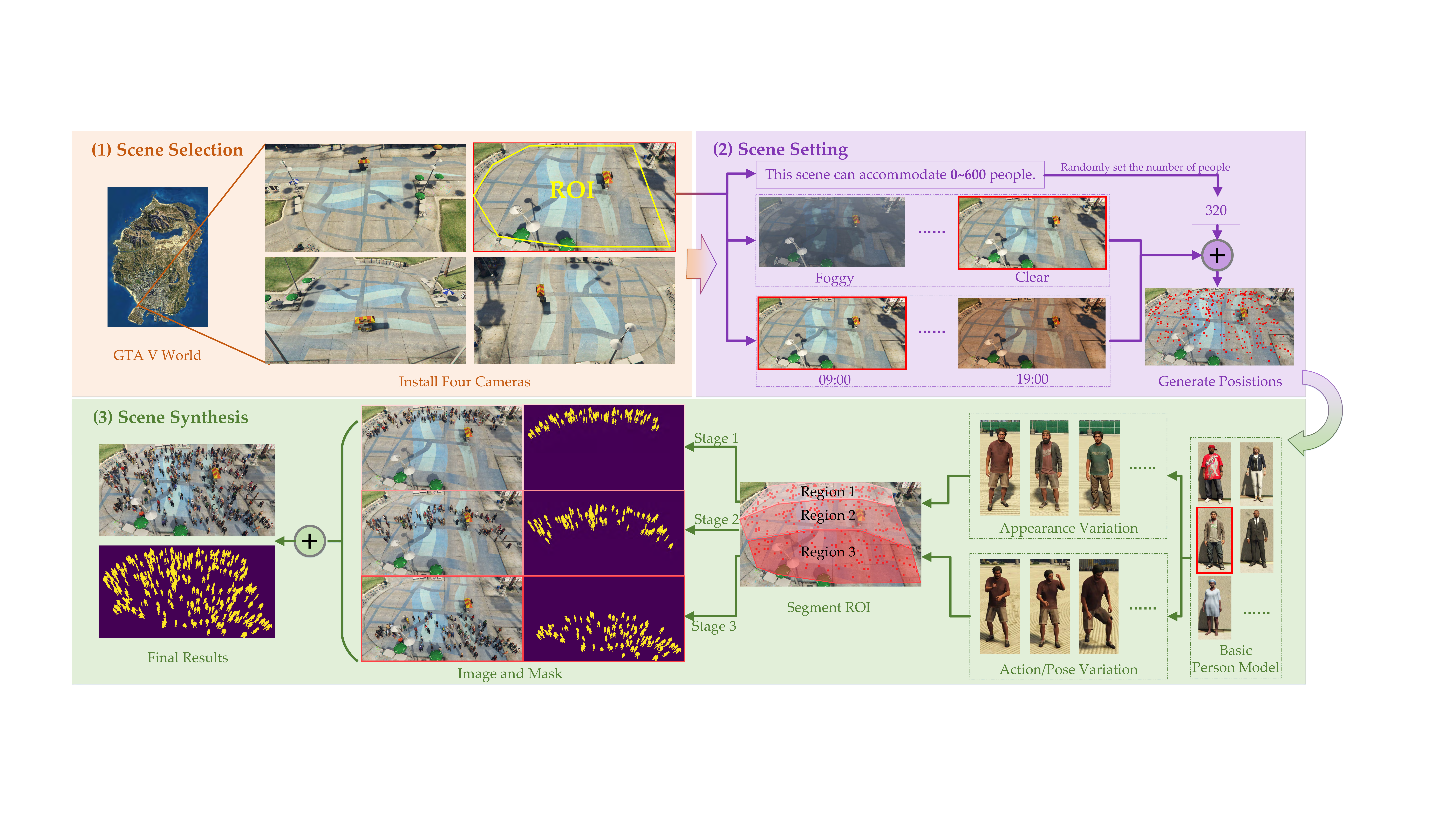}
	\caption{The illustration of crowd scene generation in GTA 5, including scene selection, setting and synthesis. }\label{Fig-flowchart}
\end{figure*}

\subsection{Synthetic Dataset}

In recent years, the mainstream deep-learning-based methods achieve a remarkable improvement relying on a large amount of training data \citep{deng2009imagenet,lin2014microsoft,abu2016youtube}. But annotating the groundtruth of massive amounts of data is a time-consuming and labor-intensive assignment, especially for pixel-wise tasks (such as semantic segmentation, density map estimation). According to the statistics, annotating a crowd scene with 1,000 people takes more than 40 minutes. To alleviate this problem, many methods about the generation and low-cost annotation of synthetic data are proposed. \cite{Richter_2016_ECCV} collect $\sim 25,000$ synthetic street scenes from GTA V. Meanwhile, they propose a low-cost annotation method that can save  90\% manual labeling time. \cite{ros2016synthia} release a synthetic semantic segmentation dataset by constructing a virtual world. They exploit Unity Engine \footnote{https://unity3d.com/} (an open-source game engine) to design various common object models, such as pedestrians, cars, buildings, etc. In the field of autonomous driving, \cite{Johnson-Roberson:2017aa} construct a synthetic auto-driving dataset from GTA V. At the same time, they present an automatic method to analyze the depth information from the game engine to get the accurate object masks. Based on GTA V, \cite{richter2017playing} propose a large-scale benchmark for autonomous driving, which provides six types of data, namely video frame, semantic segmentation mask, instance mask, optical flow, 3D layout, and visual odometry. \cite{bak2018domain} develop a synthetic person re-identification dataset based on Unreal Engine 4 \footnote{https://www.unrealengine.com/}. 

In addition to the aforementioned datasets, some famous synthetic data platforms/tools \citep{7860433,Dosovitskiy17,qiu2017unrealcv,shah2018airsim} are released. \cite{7860433} propose a test-bed platform for visual reinforcement learning, which adopts a First-Person Perspective (FPP) in the constructed semi-realistic 3D world. CARLA \citep{Dosovitskiy17} an open-source simulator for self-driving research, which supports the flexible specification of sensor suites and environmental conditions. UnrealCV \citep{qiu2017unrealcv} is a project to help researchers build virtual worlds using Unreal Engine 4, which provides some key commands to interact with the virtual world and API to connect external programs, such as Caffe \citep{jia2014caffe}. AirSim \citep{shah2018airsim} is a simulator for cars, drones or other unmanned vehicles, which is an open-source and cross platform. It supports hardware-in-loop with  flight controllers and provides depth information, RGB images and pixel-level segmentation masks.

\section{GTA5 Crowd Counting (GCC) Dataset}

\label{gcc-dataset}

Grand Theft Auto V (GTA5) is an electronic game published by Rockstar Games \footnote{https://www.rockstargames.com/} in 2013. GTA5 utilizes the proprietary Rockstar Advanced Game Engine (RAGE) to improve its draw distance rendering capabilities. Benefiting from the excellent game engine, its scene rendering, texture details, weather effects and so on are very close to the real-world conditions. Rockstar Games constructs a virtual world, including the fictional Blaine County, and the fictional city of Los Santos. GTA5 allows players freely roam the open world and explore more gameplay content. In addition, Rockstar Games allows the players to develop the mod to achieve specific needs in the game. It must be noncommercial or personal use\footnote{https://support.rockstargames.com/articles/115009494848/PC-Single-Player-Mods}$^,$\footnote{https://support.rockstargames.com/articles/200153756/Policy-on-posting-copyrighted-Rockstar-Games-material} and not be used in an online version. 

Considering the aforementioned advantages and characteristics, we develop a data collector and labeler to construct crowd scenes in GTA5, which is based on Script Hook V \footnote{http://www.dev-c.com/gtav/scripthookv/}. Script Hook V is a C++ library for developing game plugins, which allows developers to get the game data from rendering stencil. The data collector firstly constructs the congested crowd scenes via controlling the objects (pedestrians, cars, etc.) and setting attributes (weathers, timestamp, etc.) of the virtual world. Then, by analyzing the data from rendering stencil, the labeler automatically annotates the accurate head locations of persons without any manpower.

Previous synthetic GTA5 datasets \citep{Richter_2016_ECCV,Johnson-Roberson:2017aa,richter2017playing} capture normal scenes directed by the game programming. Unfortunately, there is no congested scene in GTA5. Thus, we need to design a strategy to construct crowd scenes, which is the most obvious difference with them.

\subsection{Data Collection}

\begin{figure*}
	\centering
	\includegraphics[width=1\textwidth]{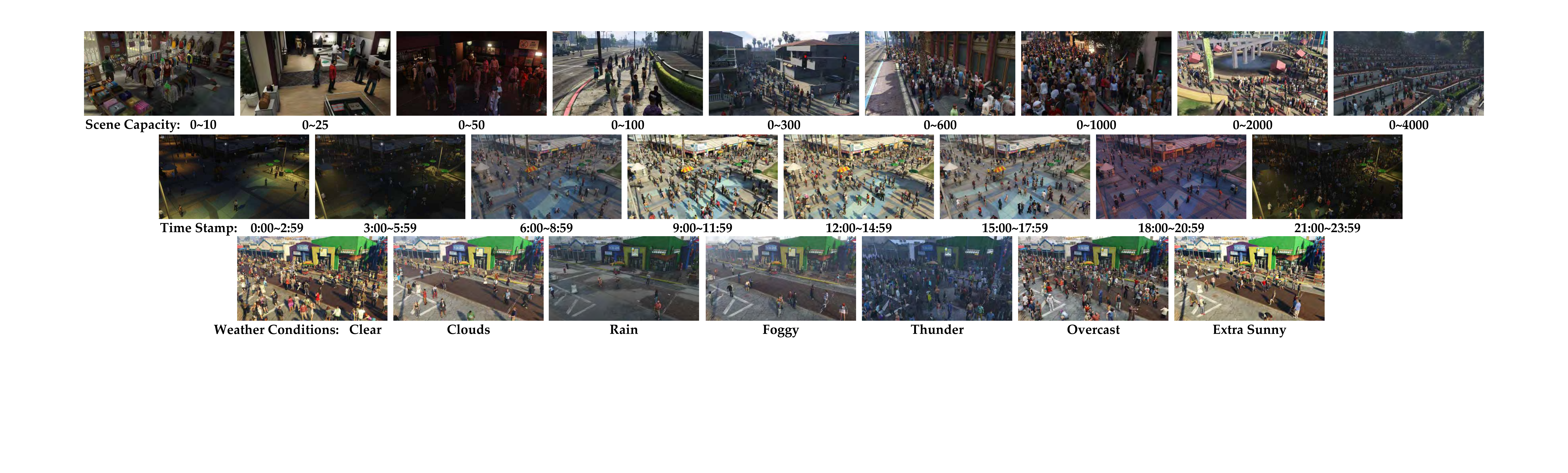}
	\caption{The display of the proposed GCC dataset from three views: scene capacity, timestamp and weather conditions. }\label{Fig-propery}
\end{figure*}

In this section, we briefly describe the key step in each component as shown in Fig. \ref{Fig-flowchart}. Scene Selection: a) select location in the GTAV world; b) equip four cameras with different appropriate parameters; c) draw a reasonable Region of interest (ROI) for crowd. Scene Setting: a) set the level of scene capacity according to the ROI's size; b) set weather conditions and time randomly; c) set the number and positions of people randomly. Scene Synthesis: a) place pedestrians in order; b) capture crowd information; c) integrate multiple scenes into one scene; d) remove the labels of occluded heads. The video demonstration is available at: \url{https://www.youtube.com/watch?v=Hvl7xWkIueo}.

\textbf{Scene Selection.} The virtual world in GTA5 is built on a fictional city, which covers an area of 252 square kilometers. In the city, we select 100 typical locations, such as the beach, stadium, mall, store and so on. To capture the scene from multiple views, the four surveillance cameras are equipped with different parameters (location, height, rotation/pitch angle) for each location. As a result, a total of 400 diverse scenes are built. In order to obtain the realistic constructed crowd scenes, we delimit a polygon area to place the person model, which is named as ``Region of Interest (ROI)''. The main purpose of this operation is to prevent people from appearing on unreasonable objects.

\textbf{Scene Setting.} After the scene selection, we need to set some basic attributes for each scene. Firstly, according to the area of ROI, we set a range of the number of people. Then, to enhance the diversity of data, the weather condition and time are randomly set to simulate various illumination and brightness during each generation. The distribution of different attributes will be reported in Section \ref{Prop}. Finally, we pre-generate the number of people and each people's position.  

\textbf{Scene Synthesis.} The last step is the scene synthesis, the most important part of the entire data construction. Due to the limitation of GTA5, the number of people can not exceed a maximum value of 256. To create a congested crowd scene, we segment several non-overlapping regions according to the distance from the Surveillance camera to and place persons in each region, as shown in Fig. \mbox{\ref{Fig-flowchart}(3)}. As for each region, we save the relevant information, such as image, segmentation mask, and head point coordinates. Next, according to the crowd mask information, all images are integrated into one scene. Finally, we remove the labels of heads occluded by other people or objects and update the label information.

The person is a core component in the crowd scene. During the scene synthesis, we employ 265 basic person models from GTAV to simulate the crowd, and every model comes with a different combination of skin color, gender, height, weight, age and so on\footnote{https://wiki.gtanet.work/index.php?title=Peds}. Besides, for each person model, it has six variations on external appearance, such as clothing, haircut, etc. Theoretically, we can create far more than person models with different appearance features. For mimicking the various poses in the real world, each person is programmed to do a random action in sparse crowd scenes.

\subsection{Properties of GCC}

\label{Prop}

GCC dataset consists of 15,212 images, with a resolution of $1080 \times 1920$, containing more than 7,000,000 persons. Compared with the existing datasets, GCC is the largest crowd counting dataset not only in the number of images but also in the number of persons. Table \ref{Table-compare} reports the basic information of GCC and the existing datasets, including data volume, image resolution, label forms and so on. In addition to the above advantages, GCC is more diverse than other real-world datasets.

\begin{figure}
	\centering
	\includegraphics[width=0.5\textwidth]{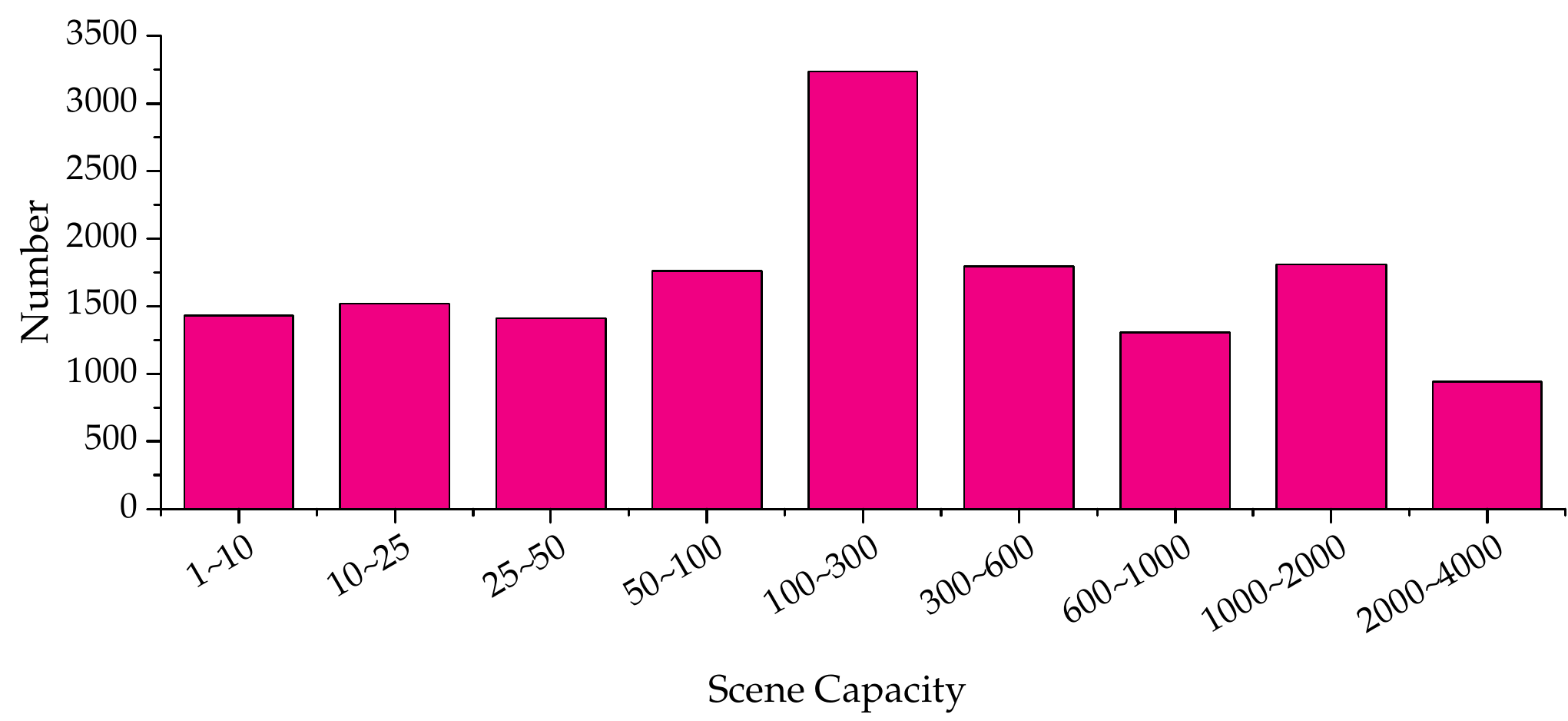}
	\caption{The statistical histogram of crowd counts on the proposed GCC dataset. }\label{Fig-hist}
\end{figure}

\textbf{Diverse Scenes.} 
In addition to the advantage in terms of data volume, GCC is more diverse than other real-world datasets. It captures 400 different crowd scenes in the GTA V game, which includes multiple types of locations. For example, indoor scenes: office, convenience store, pub, etc. outdoor scenes: mall, walking street, sidewalk, plaza, stadium and so on. Furthermore, all scenes are assigned with a level label according to their space capacity. The first row in Fig. \ref{Fig-propery} shows the typical scenes with different levels. In general, for covering the range of people, a larger scene should have more images. Thus, the setting is conducted as follows: the scenes with the first/second/last three levels contain $30/40/50$ images. Besides, these images containing some improper events should be deleted. Finally, the number of images in some scenes may be less than their expected value. Fig. \ref{Fig-hist} demonstrates the population distribution histogram of our GCC dataset. 

\begin{figure} 
	\centering 
	\subfigure[Time stamp distribution.] { \label{Fig-sector-time} 
		\includegraphics[width=0.46\columnwidth]{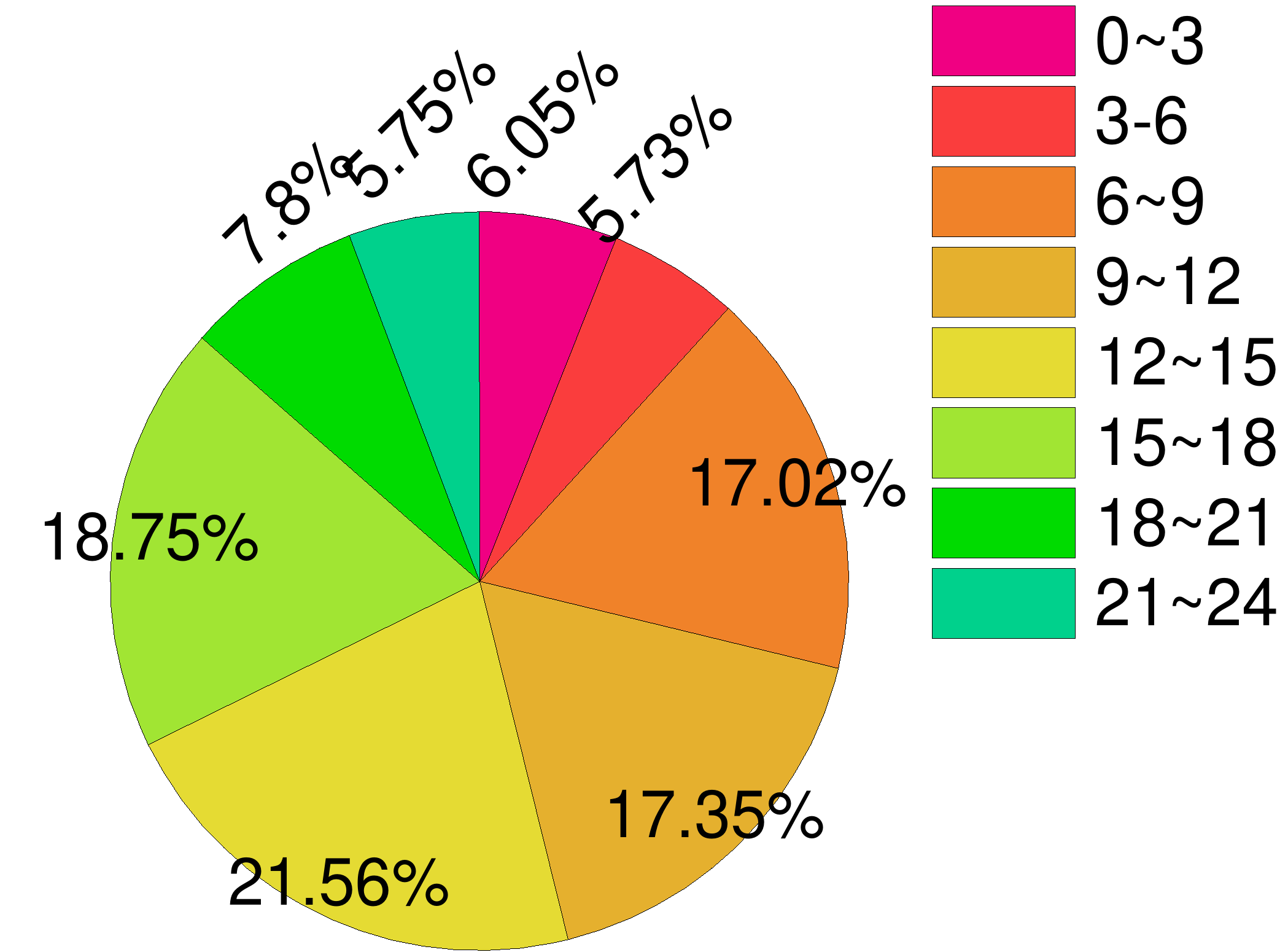}
	} 
	\subfigure[Weather condition distribution.] { \label{Fig-sector-weather} 
		\includegraphics[width=0.46\columnwidth]{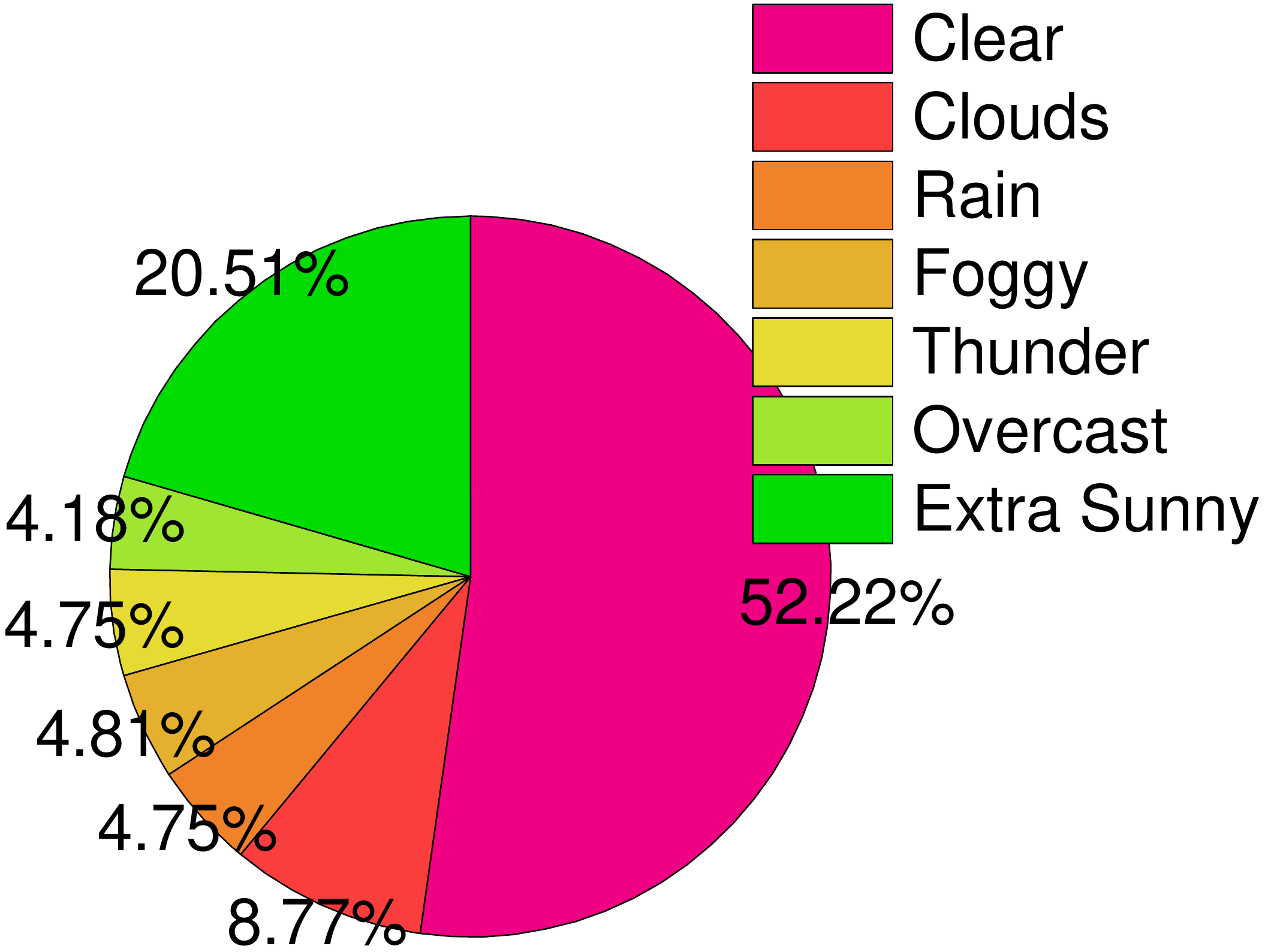}
	} 
	\caption{The pie charts of time stamp and weather condition distribution on GCC dataset. In the left pie chart, the label ``$0 \sim 3$'' denotes  the time period during $\left[ {0:00,\,\left. 3:00 \right)} \right.$ in 24 hours a day.} 
	\label{Fig-sector} 
\end{figure}


Existing datasets only focus on one of the sparse or congested crowd. However, a large scene may also contain very few people in the wild. Considering that, during the generation process of an image, the number of people is set as a random value in the range of its level. Therefore, GCC has more large-range than other real datasets. 

\textbf{Diverse Environments.} In order to construct the data that are close to the wild, the images are captured at a random time in a day and under a random weather condition. In GTA5, we select common weathers, namely clear, clouds, rain, foggy, thunder, overcast and extra sunny. The last two rows of Fig. \ref{Fig-propery} illustrate the exemplars at different times and under various weathers. In the process of generation, we tend to produce more images under common conditions. Specifically, we prefer to generate more daytime scenes and fine-weather scenes. The two sector charts in Fig. \ref{Fig-sector} respectively shows the proportional distribution on the time stamp and weather conditions of the GCC dataset. 

\textbf{Splitting for Evaluation.} In order to fully verify the performance of the algorithm, we propose three different schemes to split the dataset into two parts (namely training and testing): 
\begin{enumerate}
	\item[1)] \textbf{Random splitting}: the entire dataset is randomly divided into two groups as the training set (75\%) and testing set (25\%), respectively.
	\item[2)] \textbf{Cross-camera splitting}: as for a specific location, one surveillance camera is randomly selected for testing and the others for training.
	\item[3)] \textbf{Cross-location splitting}: we randomly choose 75/25 locations for training/testing, which can be treated as a cross-scene evaluation.
\end{enumerate}
Obviously, the task difficulty of the three strategies is increased sequentially. The last two splitting methods can effectively evaluate the generalization ability of models.

\section{Supervised Crowd Understanding}


In this section, we propose a Spatial FCN for crowd understanding, focusing on counting and segmentation tasks. 

\label{SCC}

\subsection{Spatial FCN for Crowd Understanding}



In 2014, \mbox{\cite{long2015fully}} and \mbox{\cite{kang2014fully}} propose the Fully Convolutional Network (FCN) almost simultaneously, which focuses on image segmentation. It uses the convolutional layer to replace the fully connected layer so that it can process the image with an arbitrary size and output the map of the corresponding size. To encode the large-range contextual information, \mbox{\cite{pan2017spatial}} present a spatial encoder via a sequence of convolution on the four directions (down, up, left-to-right and right-to-left).

Inspired by \cite{pan2017spatial}, we design a spatial FCN (SFCN) to estimate the crowd density maps. The spatial encoder is added to the top of the backbones: such as VGG-16 Network \mbox{\citep{simonyan2014very}} and ResNet-101 \mbox{\citep{he2016deep}}. After the spatial encoder, a regression layer is added for crowd counting, which directly outputs the density map with input's $1/8$ size. For predicting a finer crowd mask, we adopt de-convolutional layers to produce the mask with the original input's size. In addition, classification and soft-max layers are also added to the top of the network. The detailed network configurations are shown in supplementary materials. 

\noindent\textbf{Loss Function\,\,\,\,\,} Standard pixel-wise Mean Squared Error is used to optimize the proposed SFCN for crowd Counting. During the training process of the segmentation model, the objective is standard 2-D Cross Entropy loss.

\begin{figure}
	\centering
	\includegraphics[width=0.45\textwidth]{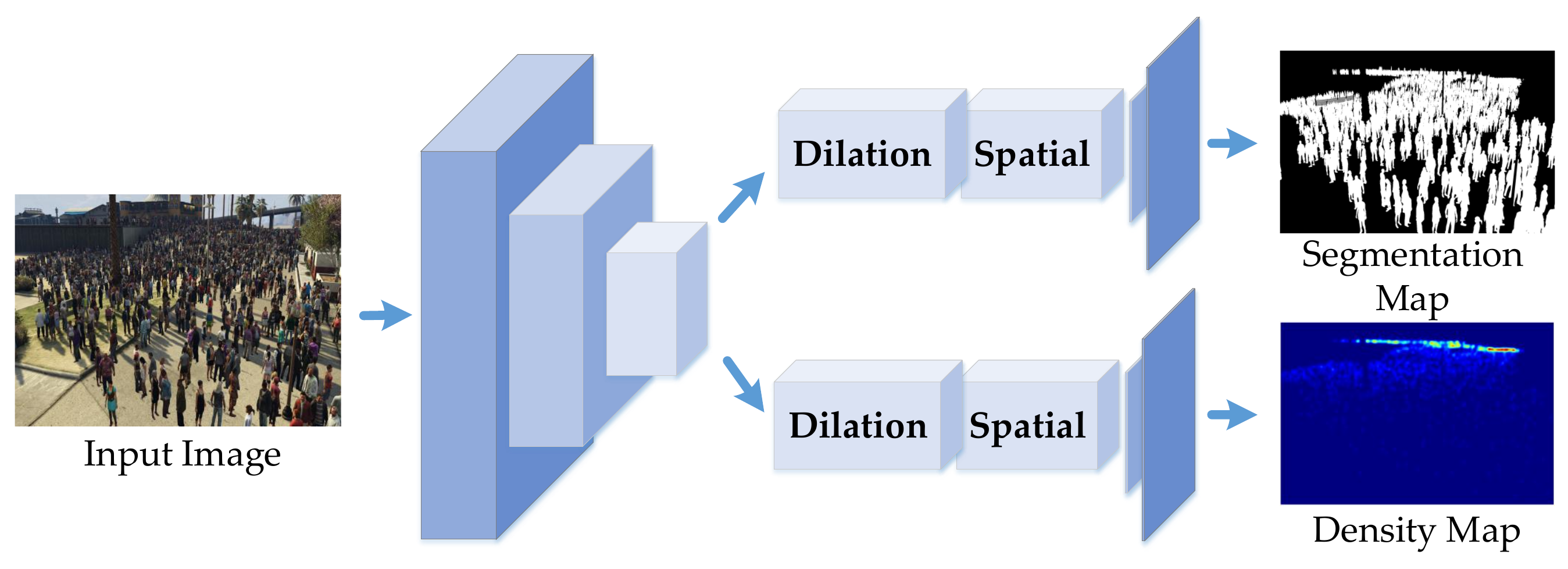}
	\caption{The architecture of spatial FCN (SFCN). }\label{Fig-sfcn}
\end{figure}

\begin{figure*}
	\centering
	\includegraphics[width=1\textwidth]{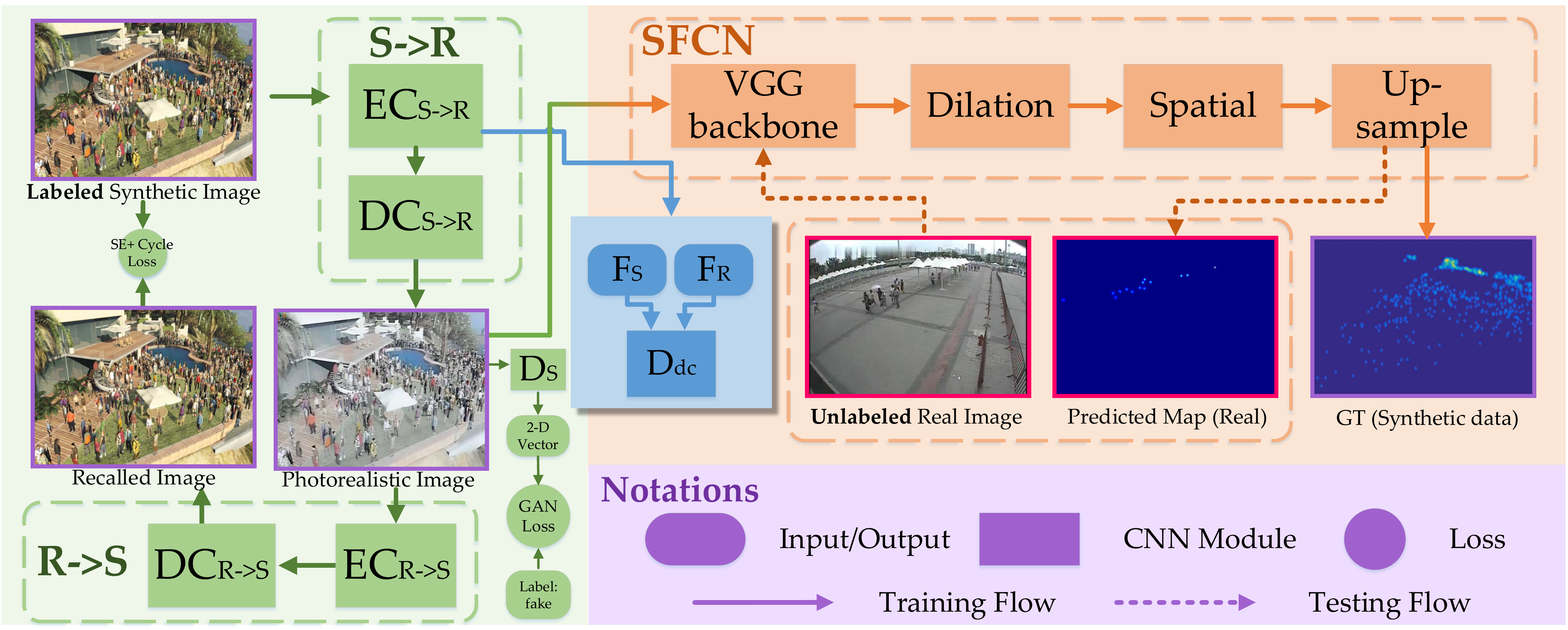}
	\caption{The flowchart of the proposed crowd counting via domain adaptation. The light green/blue regions are SSIM Embedding (SE) CycleGAN, and light orange region represents Spatial FCN (FCN). Limited by paper length, we do not show the adaptation flowchart of real images to synthetic images (R$\rightarrow$S), which is similar to S$\rightarrow$R. }\label{Fig-cyclegan}
\end{figure*}

\subsection{Pretraining \& Fine-tuning}

\label{pregcc}

Many current methods suffer from the over-fitting because of scarce real labeled data. Some methods \citep{babu2018divide,shi2018crowd,idrees2018composition} exploit the pre-trained model based on ImageNet Database \citep{deng2009imagenet} as the backbone. However, the pre-trained classification models(VGG \citep{simonyan2014very}, ResNet \citep{he2016deep} and DenseNet \citep{huang2017densely}) may be not the best selection, because they only provide the initialization for the backbone. Some specific layers (such as regression/classification modules, context encoders) are still initialized at a random or regular distributions. To remedy this problem, we propose a new pre-trained scheme, named as ``Pre-GCC'' (similarly, the traditional scheme is named as ``Pre-ImgNt''). To be specific, the designed model is firstly pre-trained on the large-scale GCC Dataset, then the pre-trained model is fine-tuned using the real data. Compared with the pre-trained model on ImageNet, Pre-GCC scheme provides entire initialized parameters so that it can alleviate over-fitting more effectively. 

\section{Crowd Understanding via Domain Adaptation}

\label{DA}

The last section proposes the Pre-GCC scheme that can significantly improve the model's performance on the real data. However, this strategy still relies on the labels of real datasets. In Section \mbox{\ref{intro}}, we have mentioned that manually annotating extremely congested scenes is a tedious task. Spontaneously, we attempt to find a new way to get rid of the burden of labeling data. Therefore, we propose a crowd counting method via Domain Adaptation (DA) to save human resources. The purpose of CU via DA is to learn the translation and domain-invariant features mapping between the synthetic domain ${\mathcal{S}}$ and the real-world domain ${\mathcal{R}}$. The synthetic domain ${\mathcal{S}}$ provides images ${I_\mathcal{S}}$ and count labels $L_\mathcal{S}$. The real-world domain ${\mathcal{R}}$ only provides images ${I_\mathcal{R}}$. In a word, given ${{\rm{i}}_\mathcal{S} \in I_\mathcal{S}}$,  ${\rm{l}}_\mathcal{S} \in L_\mathcal{S}$ and ${{\rm{i}}_\mathcal{R} \in I_\mathcal{R}}$ (the lowercase letters represent the samples in the corresponding sets), we want to train a crowd counter to predict density maps of ${\mathcal{R}}$. 


\subsection{SSIM Embedding CycleGAN}

In this section, we propose a crowd counting method via domain adaptation, which can learn domain-invariant features effectively between synthetic and real data. To be specific, we present an SSIM Embedding (SE) CycleGAN to translate the synthetic image to the photo-realistic image. At the same time, we use the adversarial learning for the SFCN counter to extract domain-invariant features in a hidden space. Finally, we directly apply the model to the real data. Fig. \mbox{\ref{Fig-cyclegan}} illustrates the flowchart of the proposed method.

\subsubsection{CycleGAN}

The original CycleGAN \mbox{\citep{zhu2017unpaired}} focuses on unpaired image-to-image translation. For different two domains, we can exploit CycleGAN to handle DA problem, which can translate the synthetic images to photo-realistic images. 
As for the domain ${\mathcal{S}}$ and ${\mathcal{R}}$, we define two generator ${G_{\mathcal{S}\rightarrow\mathcal{R}}}$ and ${G_{\mathcal{R}\rightarrow\mathcal{S}}}$. The former one attempts to learn a mapping function from domain ${\mathcal{S}}$ to ${\mathcal{R}}$, and vice versa, the latter one's goal is to learn the mapping from domain ${\mathcal{R}}$ to ${\mathcal{S}}$. To regularize the training process, the cycle-consistent loss ${\mathcal{L}_{cycle}}$ is introduced. Additionally, two discriminators ${D_{\mathcal{R}}}$ and ${D_{\mathcal{S}}}$ are modeled corresponding to the ${G_{\mathcal{S}\rightarrow\mathcal{R}}}$ and ${G_{\mathcal{R}\rightarrow\mathcal{S}}}$. Specifically, ${D_{\mathcal{R}}}$ attempts to discriminate that where the images are from ($I_\mathcal{R}$ or ${G_{\mathcal{S}\rightarrow\mathcal{R}}}(I_\mathcal{S})$), and ${D_{\mathcal{S}}}$ tries to discriminate the images from $I_\mathcal{S}$ or ${G_{\mathcal{R}\rightarrow\mathcal{S}}}(I_\mathcal{R})$. For training ${D_{\mathcal{R}}}$ and ${D_{\mathcal{S}}}$, the standard adversarial loss ${\mathcal{L}_{GAN}}$ is optimized, which is proposed by \mbox{\cite{goodfellow2014generative}}. The final loss function is defined as:
\begin{equation}
\small
\begin{array}{l}
\begin{aligned}
{\mathcal{L}_{CycleGAN}}(&{G_{\mathcal{S}\rightarrow \mathcal{R}}},{G_{\mathcal{R}\rightarrow \mathcal{S}}}, {D_\mathcal{R}},{D_\mathcal{S}},I_\mathcal{S},I_\mathcal{R}) \\
& = {\mathcal{L}_{GAN}}({G_{\mathcal{S}\rightarrow \mathcal{R}}},{D_\mathcal{R}},I_\mathcal{S},I_\mathcal{R})\\
& + {\mathcal{L}_{GAN}}({G_{\mathcal{R}\rightarrow \mathcal{S}}},{D_\mathcal{S}},I_\mathcal{S},I_\mathcal{R}) \\
& + \lambda {\mathcal{L}_{cycle}}({G_{\mathcal{S}\rightarrow \mathcal{R}}},{G_{\mathcal{R}\rightarrow\mathcal{S}}},I_\mathcal{S},I_\mathcal{R}),
\end{aligned}
\end{array}
\end{equation}
where $\lambda$ is the weight of cycle-consistent loss.

\subsubsection{SSIM Embedding CycleGAN}


In the crowd scenes, the biggest differences between high-density regions and other regions (including background regions and low-density crowd) are the local patterns and texture features instead of structural head information. 
Unfortunately, in the translation from synthetic to real images, the original cycle consistency is prone to losing them, which causes that the translated images lose the detailed information and are easily distorted. 

To remedy the aforementioned problem, we propose a Structural Similarity Index (SSIM) loss in CycleGAN, which is named as ``SE CycleGAN''. It can maintain the local structured features in the original crowd scenes. SSIM is proposed by \mbox{\cite{wang2004image}} to asses the reconstruction quality in the field of image denoising, super-resolution and so on, which computes the similarity between two images in terms of local patterns (mean, variance and covariance). In particular, when the two images are identical, the SSIM value is equal to $1$. In the practice, we convert the SSIM value into the trainable form, which is defined as: 

\begin{equation}
\small
\label{eq11}
\begin{split}
& {\mathcal{L}_{SEcycle}}({G_{\mathcal{S}\rightarrow \mathcal{R}}},{G_{\mathcal{R}\rightarrow\mathcal{S}}},I_\mathcal{S},I_\mathcal{R})\\
& = {\mathbb{E}_{{i_\mathcal{S}} \sim {I_\mathcal{S}}}}[ {1-SSIM({i_\mathcal{S}},{G_{\mathcal{R}\rightarrow\mathcal{S}}}({G_{\mathcal{S}\rightarrow{R}}}({i_\mathcal{S}}))})]\\
& + {\mathbb{E}_{{i_\mathcal{R}} \sim {I_\mathcal{R}}}}[ {1-SSIM({i_\mathcal{R}},{G_{\mathcal{S}\rightarrow\mathcal{R}}}({G_{\mathcal{R}\rightarrow{S}}}({i_\mathcal{R}}))})],
\end{split}
\end{equation}
where $SSIM( \cdot , \cdot )$ is standard computation. The first input is the original image from domain $\mathcal{S}$ or $\mathcal{R}$, and the second input is the reconstructed image produced by the two generators in turns. Finally, the final objective of SE CycleGAN $\mathcal{L}_{trans}$ is the sum of $\mathcal{L}_{CycleGAN}$ and $\mathcal{L}_{SEcycle}$.

\subsubsection{Feature-level Adversarial Learning}
\label{al}
To further prompt the adaptation performance, we add the feature-level adversarial learning for the outputs of the two generators. According to the size of each layer in $G_{\mathcal{S}\rightarrow \mathcal{R}}$ and $G_{\mathcal{R}\rightarrow \mathcal{S}}$, the generator can be treated as two components: EnCoder and DeCoder (EC and DC for short, respectively). To be specific, EC contains the down-sampling operation for image and DC has up-sampling operation. For feature-level adversarial learning, a domain classifier is present to discriminate where EC's outputs ($F_{\mathcal{S}}$and $F_{\mathcal{R}}$) are from. By the adversarial learning \mbox{\citep{goodfellow2014generative}}, the encoders can extract powerful domain-invariant features to fool the classifier. Specially, the classifier is a fully convolutional network, including four convolution layers with leaky ReLU.

As for SFCN, we select the feature maps of $I_{\mathcal{S}}^{trslt}$ and $I_\mathcal{R}$ after Spatial Module as the inputs for a domain classifier $D_{dc}$. The feature maps are written as ${F_{\mathcal{S}}}$ and ${F_{\mathcal{R}}}$, respectively. Through $D_{dc}$, the $O_\mathcal{S}$ and $O_\mathcal{R}$ can be obtained. For optimizing $D_{dc}$, a 2-D pixel-wise binary cross-entropy loss is performed. In order to confuse $D_{dc}$, the inverse adversarial loss should be added into the training of SFCN, which is defined as:
\begin{equation}
\small
\begin{array}{l}
{\mathcal{L}_{adv}(F_{\mathcal{R}})} = - \sum\limits_{F_\mathcal{R} \in \mathcal{R}} {\sum\limits_{h \in H} {\sum\limits_{w \in W} {\log (p(O_\mathcal{R}))} } },
\label{adv}
\end{array}
\end{equation}
where $O_\mathcal{R}$ is 2D-channel maps with size of $H \times W$ for real feature input $F_\mathcal{R}$, $H$ and $W$ denote the height and width of the inputs, and $p( \cdot )$ is the soft-max operation for each pixel.

\subsection{Joint Training}
\label{ad_SFCN}
Finally, the joint training of SE CycleGAN and the SFCN counter is implemented by optimizing the following loss:
\begin{equation}
\small
\begin{array}{l}
\begin{aligned}
\mathcal{L}\left( I_{\mathcal{S}},L_{\mathcal{S}},I_{\mathcal{R}} \right) =  \alpha{\mathcal{L}_{cnt}}({I_{\mathcal{S}}},{L_{\mathcal{S}}}) + \beta{\mathcal{L}_{trans}} + \lambda {\mathcal{L}_{adv}}(F_{\mathcal{R}}),
\end{aligned}
\end{array}\label{all_loss}
\end{equation}
where $\mathcal{L}_{cnt}$ is the standard MSE loss on the translated synthetic domain, $L_{adv}$ is the inverse adversarial loss for features $F_{\mathcal{R}}$ from the real domain in Section \mbox{\ref{al}}. $\alpha$, $\beta$ and $\lambda$ are the weights to balance the losses.

\subsection{Scene/Density Regularization}

For a better domain adaptation from synthetic to real world, we design two strategies to facilitate the DA model to learn domain-invariant feature and produce the valid density map.\label{SDR}

\textbf{Scene Regularization.} Since GCC is a large-counter-range and diverse dataset, using all images may cause the side effect in domain adaptation. For example, ShanghaiTech does not contain the thunder/rain scenes, and WorldExpo'10 does not have a scene that can accommodate more than 500 people. Training all translated synthetic images can decrease the adaptation performance on the specific dataset. Thus, we manually select some specific scenes for different datasets. The concrete strategies are described in Section \ref{para_da}. In general, it is a coarse data filter, not an elaborate selection. 

\textbf{Density Regularization.} Although we translate synthetic images to photo-realistic images, some objects and data distributions in the real world are unseen during training the translated images. As a pixel-wise regression problem, the density may be an arbitrary value in theory. In fact, in some preliminary experiments, we find some backgrounds in real data are estimated as some exceptionally large values. To handle this problem, we set a upper bound $MAX_{\mathcal{S}}$, which is defined as the max density in the synthetic data. If the output value of a pixel is more than $MAX_{\mathcal{S}}$, the output will be set as $0$. Note that the network's last layer is ReLU, so the output of each pixel must be greater than or equal to $0$.


\section{Experiments}
\label{exp}

\subsection{Metrics}

In the field of crowd counting, the mainstream evaluation metrics are Mean Absolute Error (MAE) and Mean Squared Error (MSE), which are formulated as follows:
\begin{equation}
\small
\begin{array}{l}
\begin{aligned}
MAE = \frac{1}{N}\sum\limits_{i = 1}^N {\left| {{y_i} - {{\hat y}_i}} \right|}, MSE = \sqrt {\frac{1}{N}\sum\limits_{i = 1}^N {{{\left| {{y_i} - {{\hat y}_i}} \right|}^2}} } ,
\end{aligned}\label{MAE}
\end{array}
\end{equation}
where $N$ is the number of samples in testing data, ${{y_i}}$ is the count label (real number of people in an image) and ${{{\hat y}_i}}$ is the estimated count value for the $i$th test sample. In addition to the evaluation of final count, we also evaluate the quality of density maps using two mainstream criteria in image assessment: Peak Signal-to-Noise Ratio (PSNR) ans Structural Similarity in Image (SSIM) \citep{wang2004image}.

For crowd segmentation task, we use Intersection-over-Union (IoU) \citep{everingham2015pascal} for crowd and background to evaluate crowd models, which is defined as:
\begin{equation}
\small
\begin{array}{l}
\begin{aligned}
IoU = \dfrac{{TP}}{{TP + FP + FN}},
\end{aligned}\label{IoU}.
\end{array}
\end{equation}
where TP, FP and FN are the numbers of true positive, false positive, and false negative samples, respectively.

\subsection{Results of Supervised Crowd Understanding}

In this section, the two types of experiments are conducted: 1) training and testing within GCC dataset; 2) pre-training on GCC and fine-tuning on the real datasets.

\subsubsection{Implementation Details}
\label{para_sc}
We use $C^3F$ \citep{gao2019c} to conduct our designed experiments, which is an open-source PyTorch \citep{paszke2017pytorch} code framework for crowd counting, and all experiments are performed on NVIDIA GTX 1080Ti GPU. Different from $C^3F$, we randomly select 10\% training data as the validation set to find the best model (which may result in some performance degradation compared with $C^3F$). As for different networks, the key hyper-parameters are listed in Table \ref{Table-training-para}.  In it, ``lr'' denotes learning rate; ``dr'' is decay rate of learning rate every epoch; ``lnf'' is short for ``label normalization factor'', which means that the density map is multiplied by a factor\footnote{This trick effectively improves the counting performance \citep{gao2019c}}. Adam \citep{kingma2014adam} algorithm is adopted to optimize the network and obtain the best result. 

In this section, the experiments involve five networks: \mbox{MCNN\citep{zhang2016single}}, \mbox{CSRNet\citep{li2018csrnet}}, FCN, SFCN and SFCN\dag. The first two are the original version in their published papers. The last three's detailed configurations are shown in Section 2.1 of the supplementary materials.

Other training parameters are logged in $C^3F$'s repository\footnote{https://github.com/gjy3035/C-3-Framework/tree/python3.x/results\\\_reports}. By flexible design of $C^3F$, our every result can be effectively reproduced. 

\begin{table}[htbp]
    	
	\centering
	\caption{The key parameters of training different models.   }
	
	\begin{tabular}{c|c|ccc}
		\whline
		Method	&backbone  &lr &dr & lnf \\
		\whline
		MCNN   &None &$10^{-4}$ &1 &100 \\
		\hline
		CSRNet & VGG-16 &$10^{-5}$ &0.995 &100\\
		\whline
		FCN & VGG-16 &$10^{-5}$ &0.995 & 100\\
		\hline	
		SFCN & VGG-16 &$10^{-5}$ &0.995 &100\\
		SFCN\dag & ResNet-101 &$10^{-5}$ &0.995 &100\\
		\whline
	\end{tabular}
	\label{Table-training-para}
\end{table}

\subsubsection{Experiments on GCC Dataset}
\label{trainGCC}
\textbf{Performance of Overall Evaluation \,\,\,\,\,} 

We report the results of the extensive experiments within the GCC dataset, which verifies SFCN from three different training strategies: random, cross-camera and cross-location splitting.  Table \ref{Table-spvsd} reports the performance of our SFCN and two popular methods on the proposed GCC dataset. In the table, ``fg'' and ``bg'' respectively denotes the foreground and background in the scenes, and ``mIoU'' is the mean value of two classes of IoU.

\begin{table}[htbp]
	\centering
	\caption{The results (MAE$\downarrow$/MSE$\downarrow$/PSNR$\uparrow$/SSIM$\uparrow$/IoU$\uparrow$) of our proposed SFCN and the two classic methods (MCNN \citep{zhang2016single} and CSRNet \citep{li2018csrnet}) on GCC dataset.}
	\small
	\setlength{\tabcolsep}{1.5mm}{
		\begin{tabular}{c|cc|cc|ccc}
			\multicolumn{8}{c}{\normalsize{Performance of random splitting}\vspace{0.1cm}}\\
			\whline
			\multirow{2}{*}{Method}	  &\multicolumn{4}{|c}{Counting} &\multicolumn{3}{|c}{Segmentation(\%)}\\
			\cline{2-8}
			&MAE &MSE &PSNR &SSIM &fg & bg &mIoU\\
			\whline
			MCNN    &100.9 &217.6 &24.00 &0.838 &77.8&40.2& 59.0\\
			\hline
			CSRNet  &38.2 &87.6 &29.52 &0.829 &94.0 &73.9 &83.9\\
			\whline
			FCN &42.3 &98.7 &30.10 &0.889&93.7 &71.2 &82.5\\
			\hline	
			SFCN &36.2 &81.1 &30.21 &0.904&94.3 &74.7 &84.5\\
			\hline	
			SFCN\dag &\textbf{28.1} &\textbf{70.2} &\textbf{31.03} &\textbf{0.927}&\textbf{94.7} &\textbf{76.1} &\textbf{85.5}\\
			\whline
		\end{tabular}
	}
	\setlength{\tabcolsep}{1.5mm}{
		\begin{tabular}{c|cc|cc|ccc}
			\multicolumn{8}{c}{\normalsize{Performance of cross-camera splitting}\vspace{0.1cm}}\\
			\whline
			\multirow{2}{*}{Method}	  &\multicolumn{4}{|c}{Counting} &\multicolumn{3}{|c}{Segmentation(\%)}\\
			\cline{2-8}
			&MAE &MSE &PSNR &SSIM &fg & bg &mIoU\\
			\whline
			MCNN    &110.0 &221.5 &23.81 &0.842 &75.5 &40.1 &57.8 \\
			\hline
			CSRNet  &61.1 &134.9 &29.03 &0.826&94.5 &73.8 &84.1 \\
			\whline
			FCN &61.5 &156.6 &28.92 &0.874&93.5 &70.7 &82.1\\
			\hline	
			SFCN &\textbf{56.0} &129.7 &29.17 &0.889&93.9 &74.3 &84.1\\
			\hline	
			SFCN\dag &57.3 &\textbf{127.3} &\textbf{30.01} &\textbf{0.895}&\textbf{94.9} &\textbf{76.6} &\textbf{85.7}\\
			\whline
		\end{tabular}
	}
	\setlength{\tabcolsep}{1.5mm}{
		\begin{tabular}{c|cc|cc|ccc}
			\multicolumn{8}{c}{\normalsize{Performance of cross-location splitting}\vspace{0.1cm}}\\
			\whline
			\multirow{2}{*}{Method}	  &\multicolumn{4}{|c}{Counting} &\multicolumn{3}{|c}{Segmentation(\%)}\\
			\cline{2-8}
			&MAE &MSE &PSNR &SSIM &fg & bg &mIoU\\
			\whline
			MCNN    &154.8 &340.7 &24.05 &0.857 &76.3 &37.4 &56.9 \\
			\hline
			CSRNet  &92.2 &220.1 &28.75 &0.842 &94.4 &73.3 &83.9\\
			\whline
			FCN &97.5 &226.8 &29.33 &0.866&93.4 &69.9 &81.7\\
			\hline	
			SFCN &89.3 &216.8 &29.50 &0.906&94.9 &73.8 &84.3\\
			\hline	
			SFCN\dag &\textbf{83.9} &\textbf{209.7} &\textbf{29.76} &\textbf{0.914}&\textbf{95.1} &\textbf{76.0} &\textbf{85.5}\\
			\whline
		\end{tabular}
	}
	\label{Table-spvsd}
\end{table}

From the table, we find SFCN\dag\, attains the best performance in both crowd counting and segmentation tasks, which is due to the more  powerful learning ability of ResNet-101 than VGG-16 Net. For a fair comparison, we select CSRNet, FCN and SFCN that use the same backbone (VGG-16) to show the effectiveness of the proposed SFCN. We find SFCN is better than CSRNet and FCN in terms of seven metrics on counting and segmentation. 

In addition, from the counting performance of the three aspects (random, cross-camera and cross-location splitting), the performances are decreased significantly, which means the difficulty of three tasks is rising in turn. The main reason is that there is a big difference in the distribution of people in different crowd scenes. In contrast, the segmentation results of different models in the three evaluations are very similar, which implies that crowd region segmentation is not sensitive to different evaluation strategies. The essential reasons are: 1) GCC's person model is fixed though the crowd scenes are different; 2) the segmentation focuses on appearance feature.

\noindent\textbf{Multi-task Learning for Counting and Segmentation  \,\,\,\,\,} 

Counting and segmentation are two complementary tasks: the former focuses on the local density, and the latter aims at the difference between foreground and background. On the one hand, introducing segmentation can effectively reduce the error density estimation in background regions. On the other hand, density maps provide rich information to represent different-density crowd regions, which aids the segmentation branch in tackling them via different priors. Here, we also conduct the experiments of multi-task learning using SFCN on GCC. During the training stage, the loss weights for counting and segmentation are $1$, $0.01$, respectively. Table \mbox{\ref{mtl}} shows the counting (MAE/MSE) and segmentation (mIoU) performance of single-task learning (STL) and multi-task learning (MTL). From the table, we find that MTL outperforms the STL in terms of counting and segmentation performance. 

\begin{table}[htbp]
	\centering
	\caption{The results of STL and MTL on GCC dataset.}
	\setlength{\tabcolsep}{2.5mm}{
		\begin{tabular}{c|c|c}
			\whline
			\multirow{2}{*}{Data}	  &Single task &Multi task\\
			\cline{2-3}
			&MAE/MSE/mIoU &MAE/MSE/mIoU\\
			\whline
			rd     &36.2/81.1/84.5 &33.9/80.6/85.7 \\
			\hline
			cc  &56.0/129.7/84.1 &52.6/125.4/84.7 \\
			\whline
			cl &89.3/216.8/84.3 &85.7/209.9/85.5 \\
			\whline	

		\end{tabular}
	}\label{mtl}
\end{table}

\subsubsection{Comparison of Different Pre-trained Models}

In Section \mbox{\ref{pregcc}}, we propose a pre-training scheme to provide a model with better-initialized parameters, which can significantly improve the performance on small-scale counting datasets. To verify our strategy, we conduct the MCNN, CSRNet, SFCN and SFCN\dag\, on the two datasets (UCF-QNRF and SHT B) and compare different pre-training data. Notably, there are five strategies: 

\textbf{FS:} train the model From Scratch (light-wight models use it, such as MCNN);

\textbf{Pre-ImgNt:} Pre-train the model on ImageNet and fine-tune on a specific dataset (mainstream VGG-backbone or ResNet-backbone models use it, such as CSRNet);

\textbf{Pre-GCC:} Pre-train the model on GCC dataset and fine-tune on a specific dataset;

\textbf{Pre-UR:} Pre-train the model on the Union of seven Real-world datasets (UCSD, Mall, UCF\_CC\_50, WorldExpo'10, SHT A, SHT B and UCF-QNRF), and fine-tune on a specific dataset;

\textbf{Pre-GU:} Pre-train the model on GCC and UR (Union of Real datasets, same as the seven aforementioned datasets), and fine-tune on a specific dataset.

\begin{table*}[htbp]
	\small
	\centering
	\caption{The fine-tuning SFCN's results (MAE/MSE) on the two real-world datasets by using three different pre-trained models: Pre-GCC, Pre-UR and Pre-GU. The \textbf{\textcolor{blue}{blue bold}} texts denote the baseline results. The relative reduction is computed based on the corresponding baseline.}
	\setlength{\tabcolsep}{1.5mm}{
		\begin{tabular}{cIc|c|c|cIc|c|c|cIc}
			\whline
			\multirow{2}{*}{Method} &	\multicolumn{4}{cI}{UCF-QNRF} & \multicolumn{4}{cI}{SHT B} & \multirow{2}{*}{\tabincell{c}{Avg.\\ Reduction}} \\
			\cline{2-9} 
			& MCNN &CSRNet & SFCN &SFCN\dag & MCNN &CSRNet & SFCN &SFCN\dag \\
			\whline
			FS &\textbf{\textcolor{blue}{281.2/445.0}} &- &- &- &\textbf{\textcolor{blue}{26.3/39.5}} &- &- &-  &- \\
			\hline
			Pre-ImgNt  &- &\textbf{\textcolor{blue}{120.3/208.5}} & \textbf{\textcolor{blue}{134.3/240.3}} & \textbf{\textcolor{blue}{114.8/192.0}}  &- & \textbf{\textcolor{blue}{10.6/16.0}} & \textbf{\textcolor{blue}{11.0/17.1}} &\textbf{\textcolor{blue}{8.9/14.3}} &-\\
			\hline
			Pre-GCC  & \tabincell{c}{199.8/311.2\\ (\textcolor{red}{$\downarrow 29/30\%$})} & \tabincell{c}{112.4/185.6\\ (\textcolor{red}{$\downarrow 7/11\%$})} & \tabincell{c}{124.7/203.5\\ (\textcolor{red}{$\downarrow 7/15\%$})} & \tabincell{c}{102.0/171.4\\ (\textcolor{red}{$\downarrow 11/11\%$})} &\tabincell{c}{18.8/28.2\\(\textcolor{red}{$\downarrow 29/29\%$})} &\tabincell{c}{10.1/15.7\\(\textcolor{red}{$\downarrow 5/2\%$})} &  \tabincell{c}{9.4/14.4\\(\textcolor{red}{$\downarrow 15/16\%$})}  &  \tabincell{c}{7.6/13.0\\(\textcolor{red}{$\downarrow 15/9\%$})} &\textcolor{red}{$\downarrow 15\%$} \\
			\hline
			Pre-UR &\tabincell{c}{194.5/304.9\\(\textcolor{red}{$\downarrow 31/31\%$})} &\tabincell{c}{115.7/180.4\\(\textcolor{red}{$\downarrow 4/13\%$})} &  \tabincell{c}{110.4/187.1\\(\textcolor{red}{$\downarrow 18/22\%$})} &\tabincell{c}{107.8/186.2\\(\textcolor{red}{$\downarrow 6/3\%$})} &\tabincell{c}{19.5/30.2\\(\textcolor{red}{$\downarrow26/24\%$})} &\tabincell{c}{10.2/16.0\\(\textcolor{red}{$\downarrow4/0\%$})} &\tabincell{c}{9.3/14.6\\(\textcolor{red}{$\downarrow15/15\%$})} &\tabincell{c}{8.3/13.6\\(\textcolor{red}{$\downarrow 7/5\%$})} &\textcolor{red}{$\downarrow 14\%$}\\
			\hline
			Pre-GU  &\tabincell{c}{177.5/271.5\\(\textcolor{red}{$\downarrow 37/39\%$})}  &\tabincell{c}{104.2/171.9\\(\textcolor{red}{$\downarrow 13/18\%$})} &  \tabincell{c}{106.6/171.9			\\(\textcolor{red}{$\downarrow 21/28\%$})} &\tabincell{c}{\textbf{98.6}/\textbf{170.2}\\(\textcolor{red}{$\downarrow 14/11\%$})} &\tabincell{c}{17.4/27.5\\(\textcolor{red}{$\downarrow 34/30\%$})} &\tabincell{c}{9.9/15.3\\(\textcolor{red}{$\downarrow 7/4\%$})} &\tabincell{c}{9.1/14.5\\(\textcolor{red}{$\downarrow 15/17\%$})} &\tabincell{c}{\textbf{7.2}/\textbf{12.4}\\(\textcolor{red}{$\downarrow 19/13\%$})} & \textcolor{red}{$\downarrow 20\%$}\\
			\whline
			
	\end{tabular}}
	\label{Table-ft-new}
\end{table*}

\begin{figure*}
	\centering 
	\subfigure[] { 
		\includegraphics[width=0.65\columnwidth]{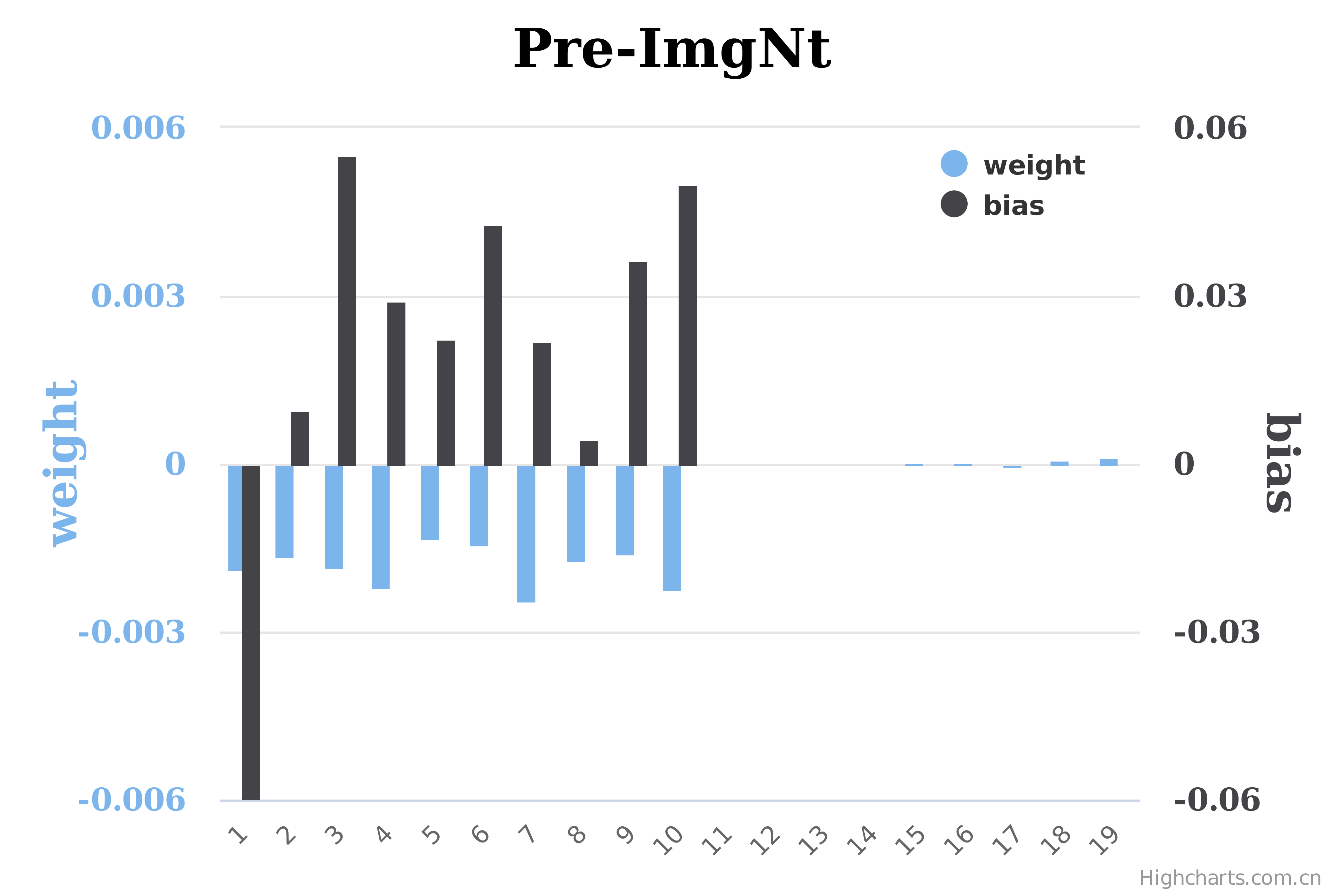}
	} 
	\subfigure[] {
		\includegraphics[width=0.65\columnwidth]{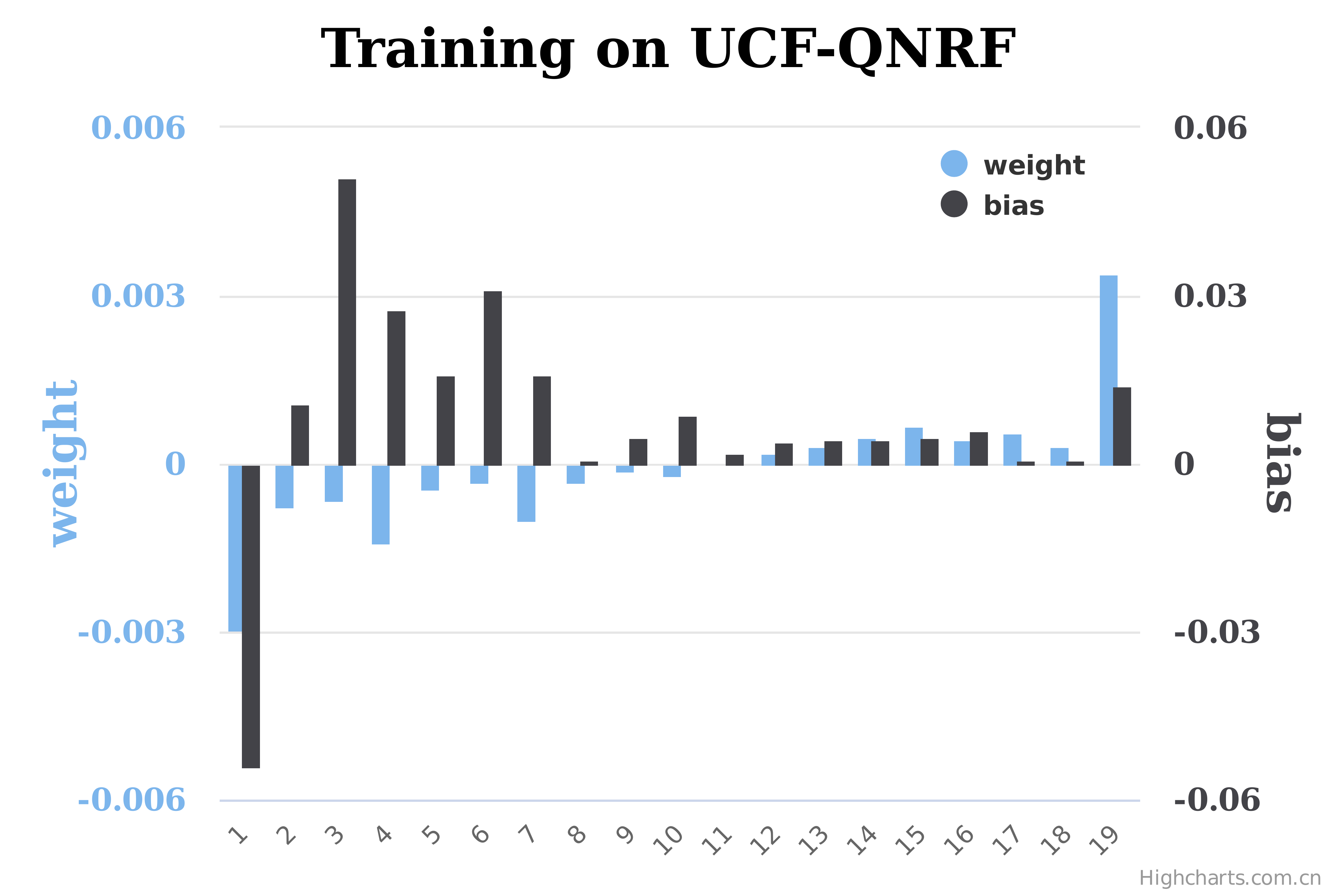}
	}
	\\ 
	\subfigure[] {
		\includegraphics[width=0.65\columnwidth]{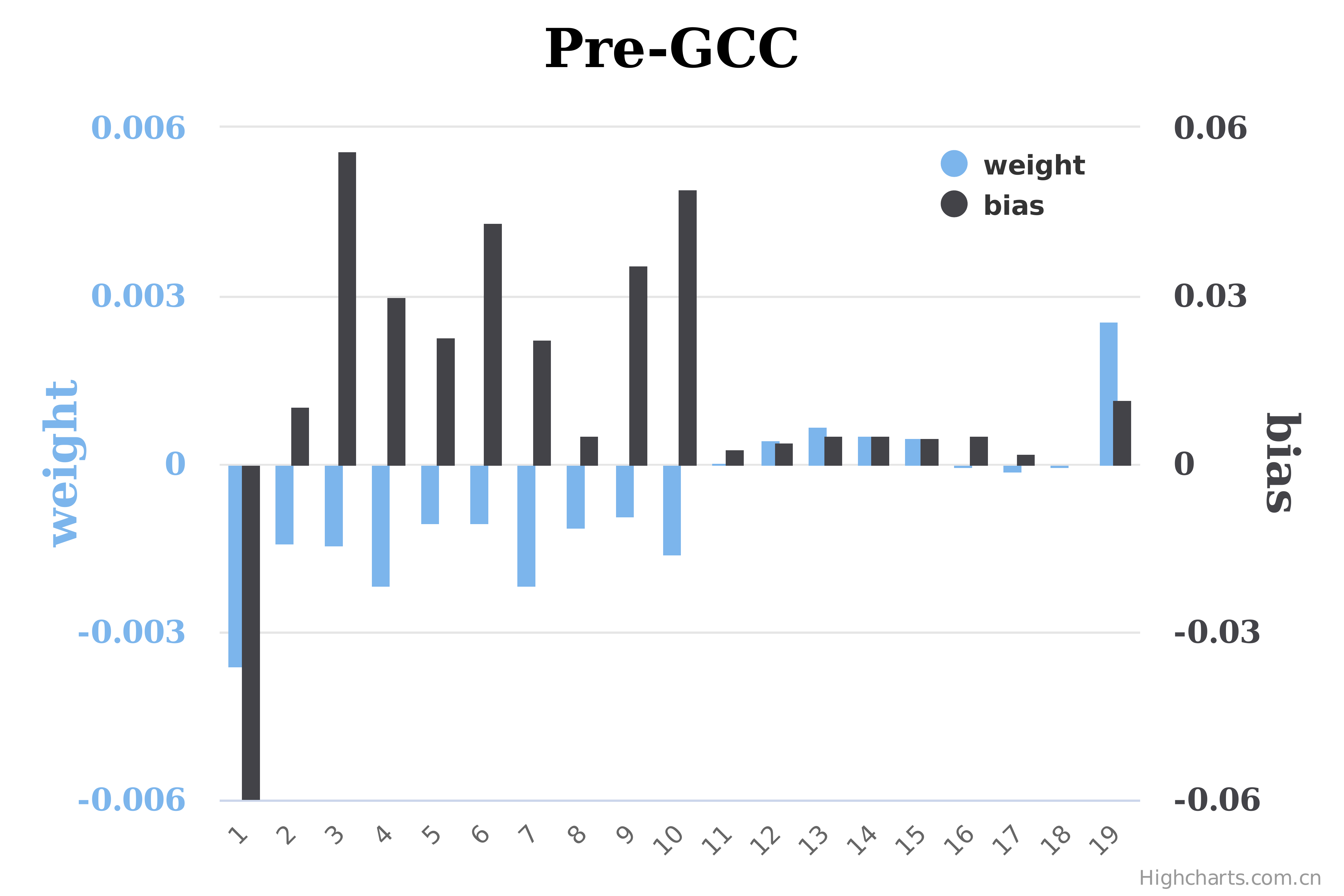}
	}
	\subfigure[] {
		\includegraphics[width=0.65\columnwidth]{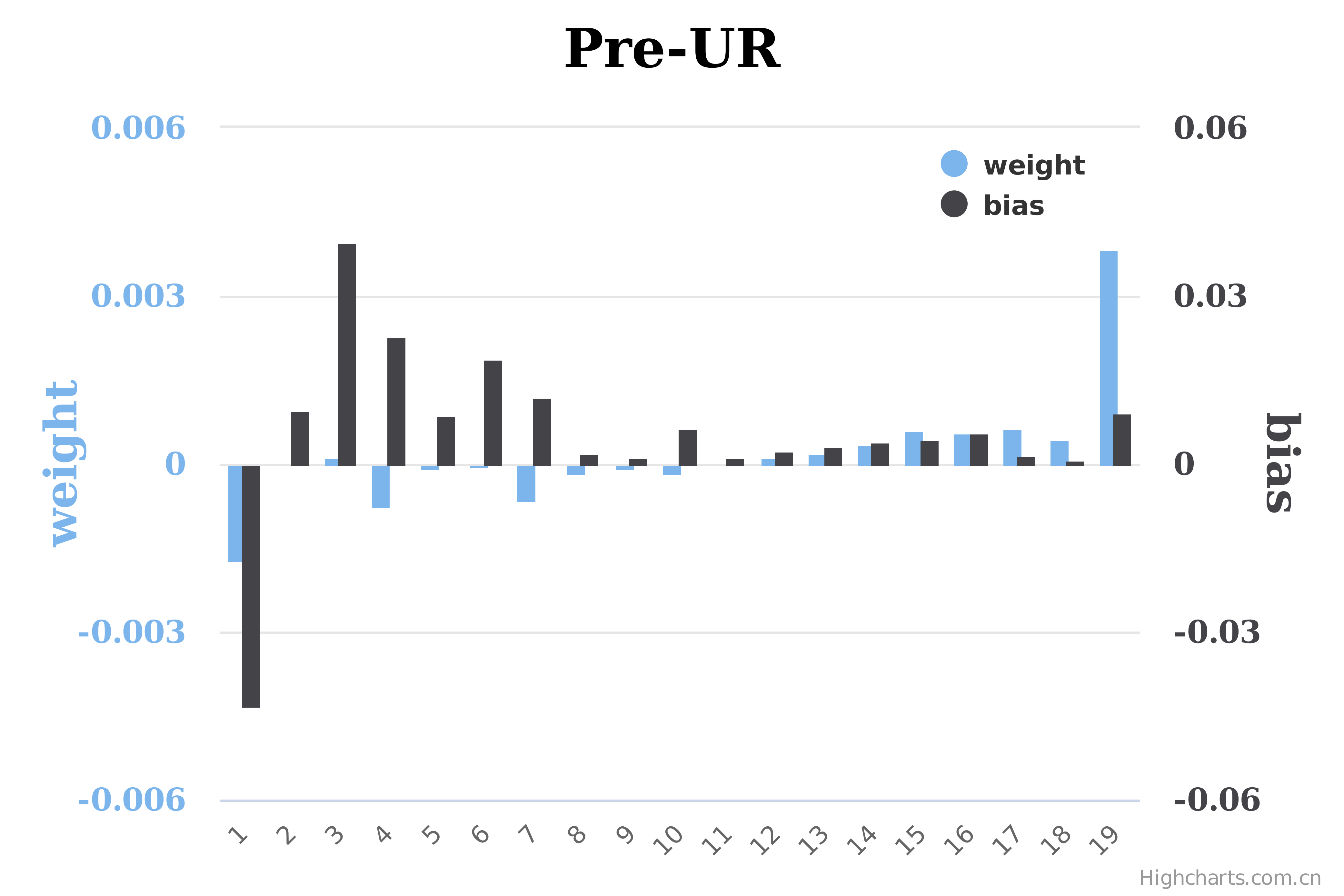}
	}
	\subfigure[] {
		\includegraphics[width=0.65\columnwidth]{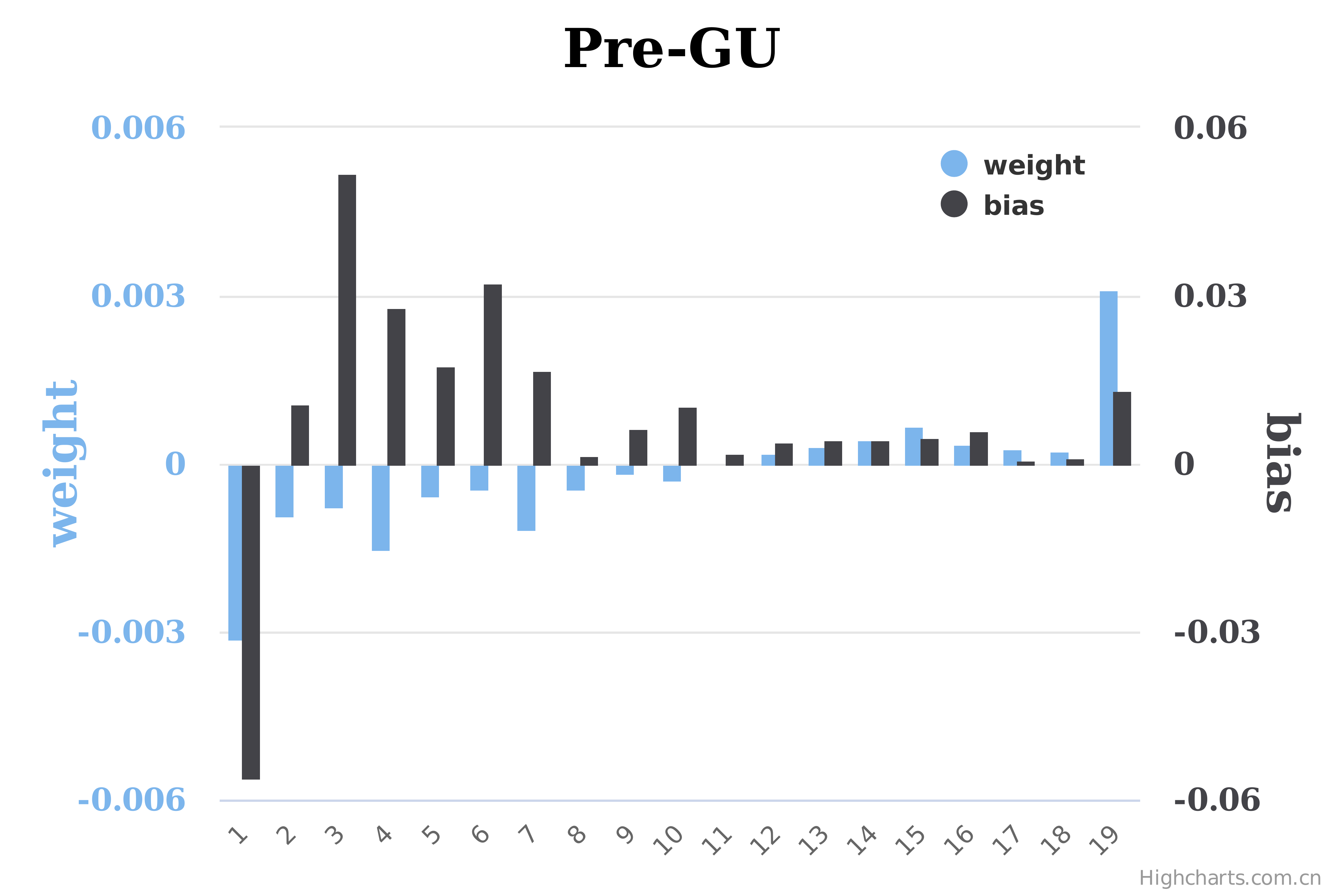}
	}
	\caption{The mean value of weight and bias in each layer of SFCN models with different training strategies. (a) Pre-ImgNt: pre-trained model on ImgNt, (b) traditional supervised training on UCF-QNRF, (c,d,e) pre-trained model on GCC, UR and GU. } 
	\label{Fig-para} 
\end{figure*}

Note that all pre-trained data are from the training set. Considering the disparity in the data volume of each dataset, each subset is sampled with the same probability during the training stage of Pre-UR and Pre-GU. In addition, small images (such as UCSD) will be resized to at least 480px. Other settings are the same as Section \mbox{\ref{para_sc}}. Table \mbox{\ref{Table-ft-new}} shows the fine-tuning SFCN's results on the two real-world datasets by using three different pre-trained models. The bold blue fonts represent the baseline results, and the red percentages indicate the relative reduction compared with the corresponding baseline. From the table, there are two interesting findings:

\begin{enumerate}
	\item[1)] Using the extra pre-trained counting data can effectively prompt the performance. The proposed Pre-GCC, Pre-UR and Pre-GU reduce the average error by 15\%, 14\% and 20\%, respectively. We also find that Pre-GU's results are better than Pre-GCC and Pre-UR. 
	\item[2)] The average performance improvement is more significant based on the model trained from Scratch (MCNN) than the model pre-trained on ImageNet (CSRNet, SFCN and SFCN\dag): $\sim 31\%$ \emph{v.s.} $\sim 12\%$.
\end{enumerate}

\begin{figure*}[htbp]
	\centering
	\includegraphics[width=0.98\textwidth]{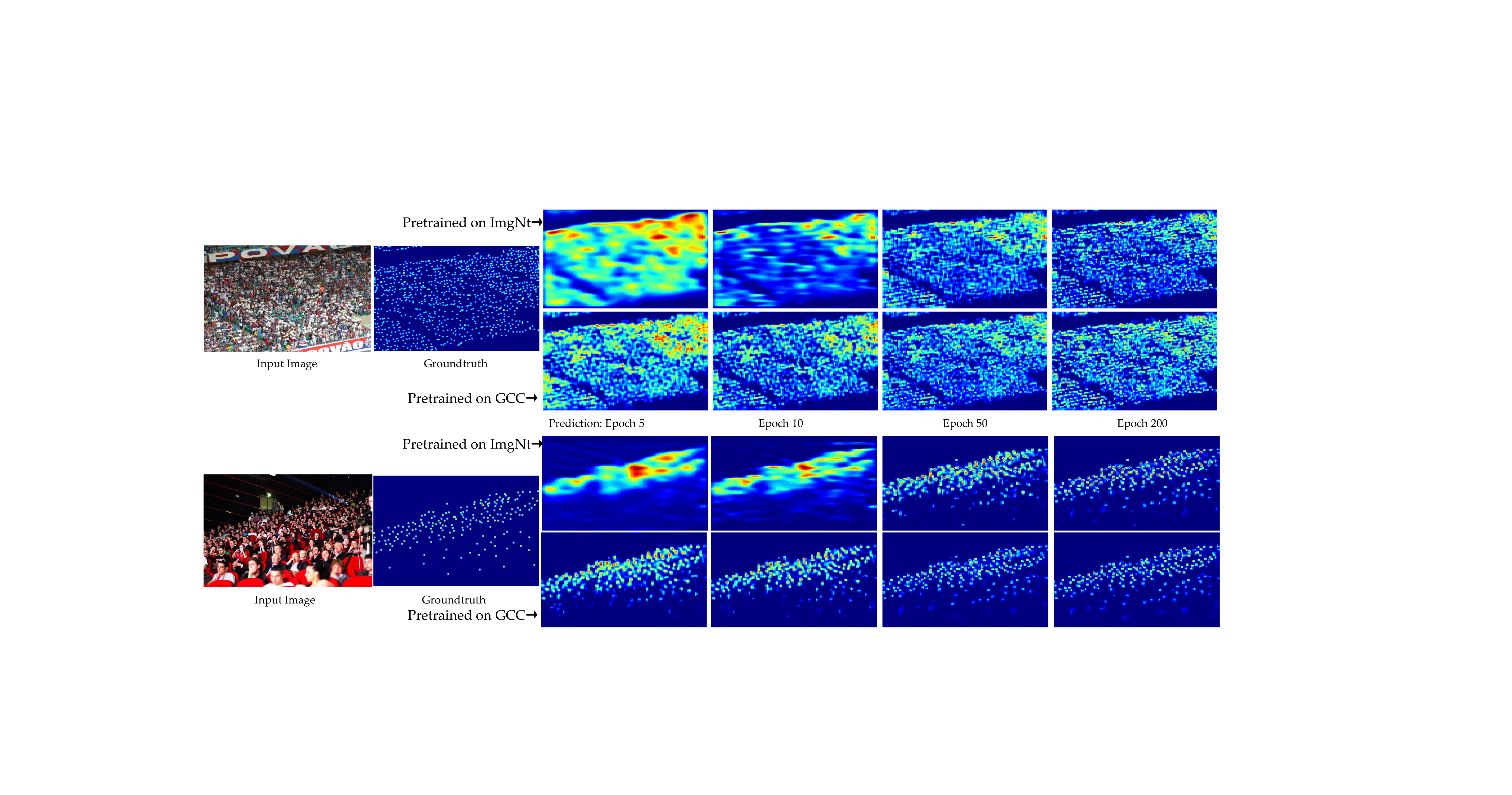}
	\caption{Visual comparison of different pre-trained models on UCF-QNRF. }\label{Fig-viscom}
\end{figure*}

\begin{figure}
	\centering
	\includegraphics[width=0.45\textwidth]{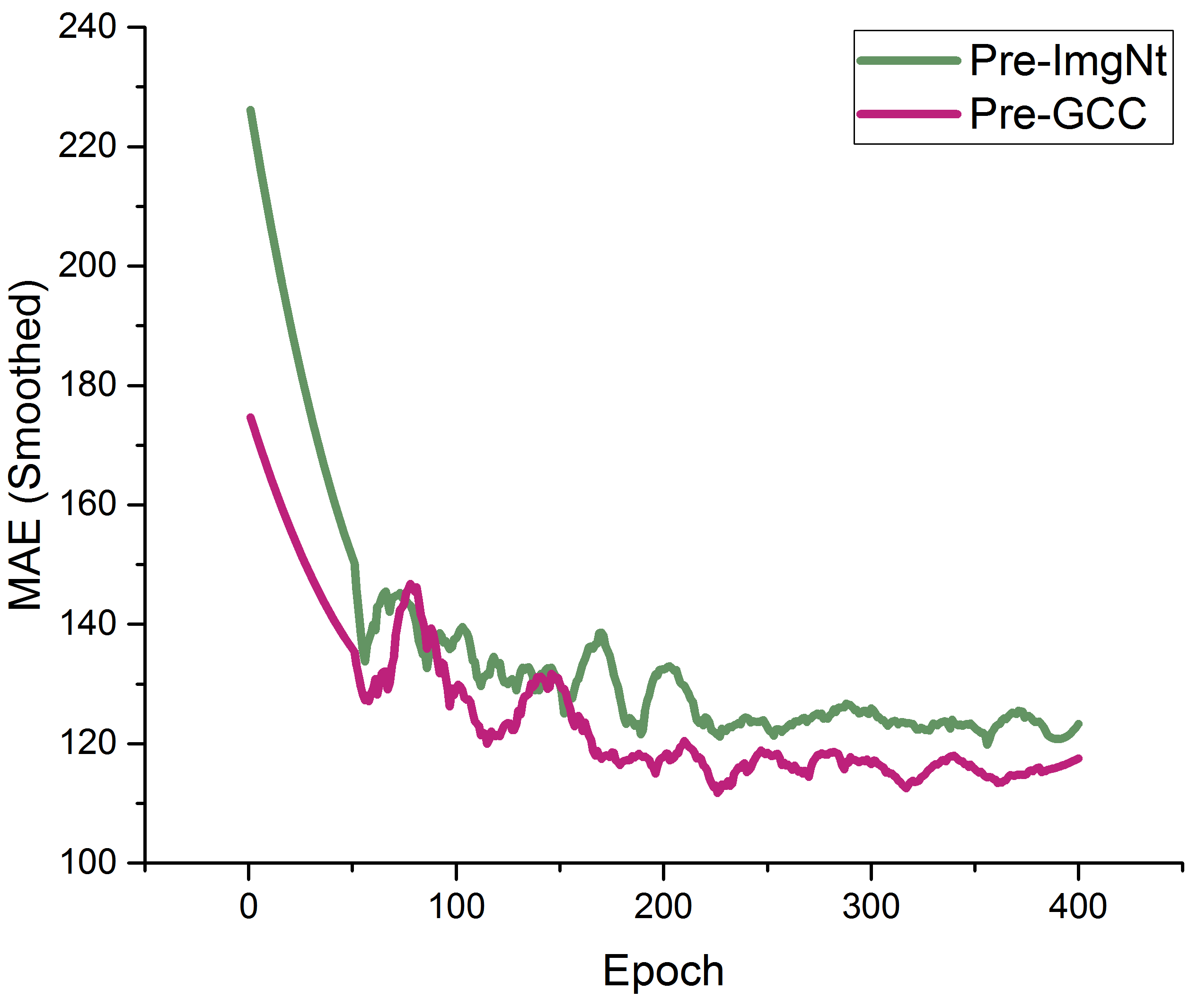}
	\caption{The MAE curve of two different training schemes on UCF-QNRF test set. }\label{Fig-loss}
\end{figure}

\vspace{0.2cm}
\noindent\textbf{Comparison on the Parameter Level}

In addition to the comparison of the final estimation results, we further explore the differences in the pre-training as mentioned above models at the parameter level. Take SFCN as an example, we compute the average distribution of weights and bias for each layer for four pre-trained models (Pre-ImgNt, Pre-GCC, Pre-UR, Pre-GU) and a fine-tuning model based on Pre-GU. Fig. \mbox{\ref{Fig-para}} illustrates the mean value of weight and bias in each layer of SFCN models with different training strategies. For Fig. \mbox{\ref{Fig-para}}(a), the first ten layers are VGG-16 backbone, and the others are randomly initialized. By comparing Fig. \mbox{\ref{Fig-para}}(a) and (b), the distribution difference between the two is very obvious. However, other pre-trained models' distributions on counting datasets (namely Pre-GCC, UR and GU) are very close to Fig. \mbox{\ref{Fig-para}}(b). The similarity of the last four models shows that crowd counting models have a certain distribution. Introducing pre-training scheme on counting data provides better installation parameters than training from scratch or pre-training from other tasks. Besides, we also find Pre-UR and Pre-GU are more similar to Fig. \mbox{\ref{Fig-para}}(b) than Pre-GCC. The main reason is that the first two pre-training methods use the labeled UCF-QNRF data.

\vspace{0.2cm}
\noindent\textbf{Visual Comparison of Different Pre-trained Models}

The last section shows the improvement of performance after using Pre-GCC. In order to show this effectiveness, we report the MAE curve and different visual density maps of two pre-trained strategies (namely pre-trained on ImgNt and GCC) during the entire training process. Fig. \ref{Fig-loss} depicts the variation curve of the loss during the training process. At the beginning of training, the model using Pre-GCC can converge rapidly than the model using Pre-ImgNt, which means that the initialized parameter provided by the former is better the latter. Besides, we find the purple line is lower than the green curve. In other words, exploiting the Pre-GCC strategy can achieve better training results. For an intuitive comparison, we record the visualization results of density map estimation during different training phases, which are shown in Fig. \ref{Fig-viscom}. To be specific, we select two typical test images and show their density map at Epoch 5, 10, 50 and 200 in the entire training stage. In the early stages of training (before Epoch 10), Pre-GCC can easily get an acceptable result, while the Pre-ImgNt can only output a coarse density distribution. This phenomenon confirms the convergence curve in Fig.\ref{Fig-loss}. As the training continues, Pre-ImgNt can also output a fine density map after Epoch 200, though the result is worse than Pre-GCC.

\vspace{0.2cm}
\noindent\textbf{Generalization Ability of Different Pre-trained Models }

Here, we further compare the cross-dataset generalization ability of counters using different pre-trained models. To be specific, we apply the two SFCN\dag \, models trained on UCF-QNRF dataset using different training schemes to the two other real-world counting dataset, namely SHT A and B. Table \ref{Table-gen} shows the experimental results. From it, we find Pre-GCC can significantly prompt the model's generalization ability. Specifically, compared with Pre-ImgNt, Pre-GCC can reduce the MAE by 6.1\% ($108.0\rightarrow101.4$) and 17.4\% ($17.2\rightarrow14.2$) on SHT A and B, respectively. The better generalization ability means that Pre-GCC makes the model perform better in unseen real data than the traditional Pre-ImgNt.

\begin{table}[htbp]
	
	\centering
	\caption{The performance (MAE/MSE) of different UCF-QNRF counting models (SFCN\dag) under the different pre-trained models on SHT A and B dataset.}
	
	\begin{tabular}{c|c|c}
		\whline
		Strategy &SHT A &SHT B\\
		\whline
		Pre-ImgNt  &108.0/184.1  &17.2/24.9  \\
		\hline
		Pre-GCC   &\tabincell{c}{\textbf{101.4}/\textbf{179.2}\\(\textcolor{red}{$\downarrow 6.1/2.7\%$})} &\tabincell{c}{\textbf{14.2}/\textbf{21.4}\\(\textcolor{red}{$\downarrow 17.4/14.1\%$})} \\
		\whline
		
	\end{tabular}
	\label{Table-gen}
\end{table}


\subsubsection{The Effect of Pre-GCC on the Fine-tuning Results}

In this section, we analyze how different GCC data affect the final fine-tuning result on the real-world datasets (UCF-QNRF and SHT B), such as different data volumes, different combinations of scene levels, and different-luminance crowd scenes. 

\begin{figure} [h]
	\centering 
	\subfigure[UCF-QNRF.] { 
		\includegraphics[width=0.45\columnwidth]{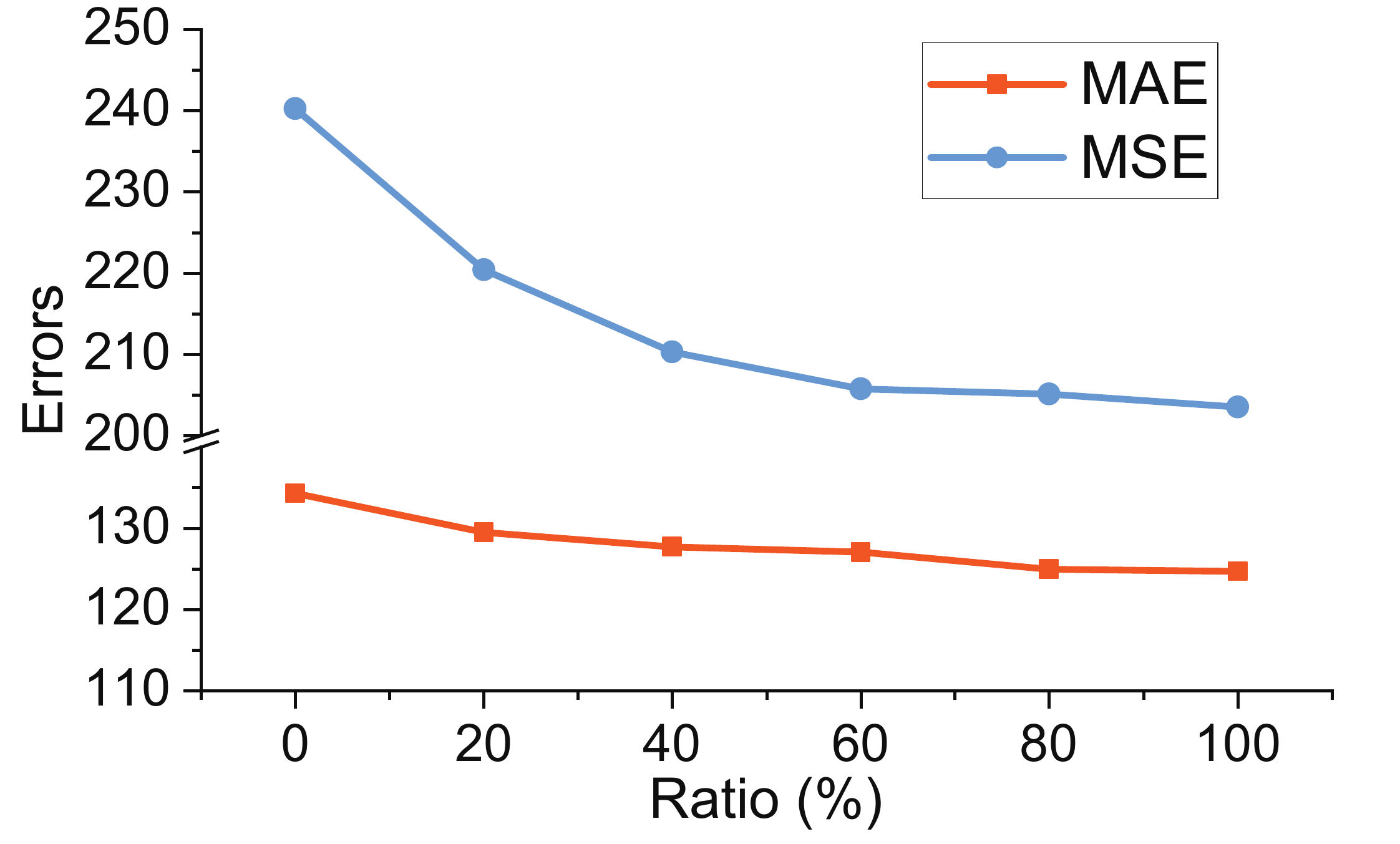}
	} 
	\subfigure[SHT B.] {
		\includegraphics[width=0.45\columnwidth]{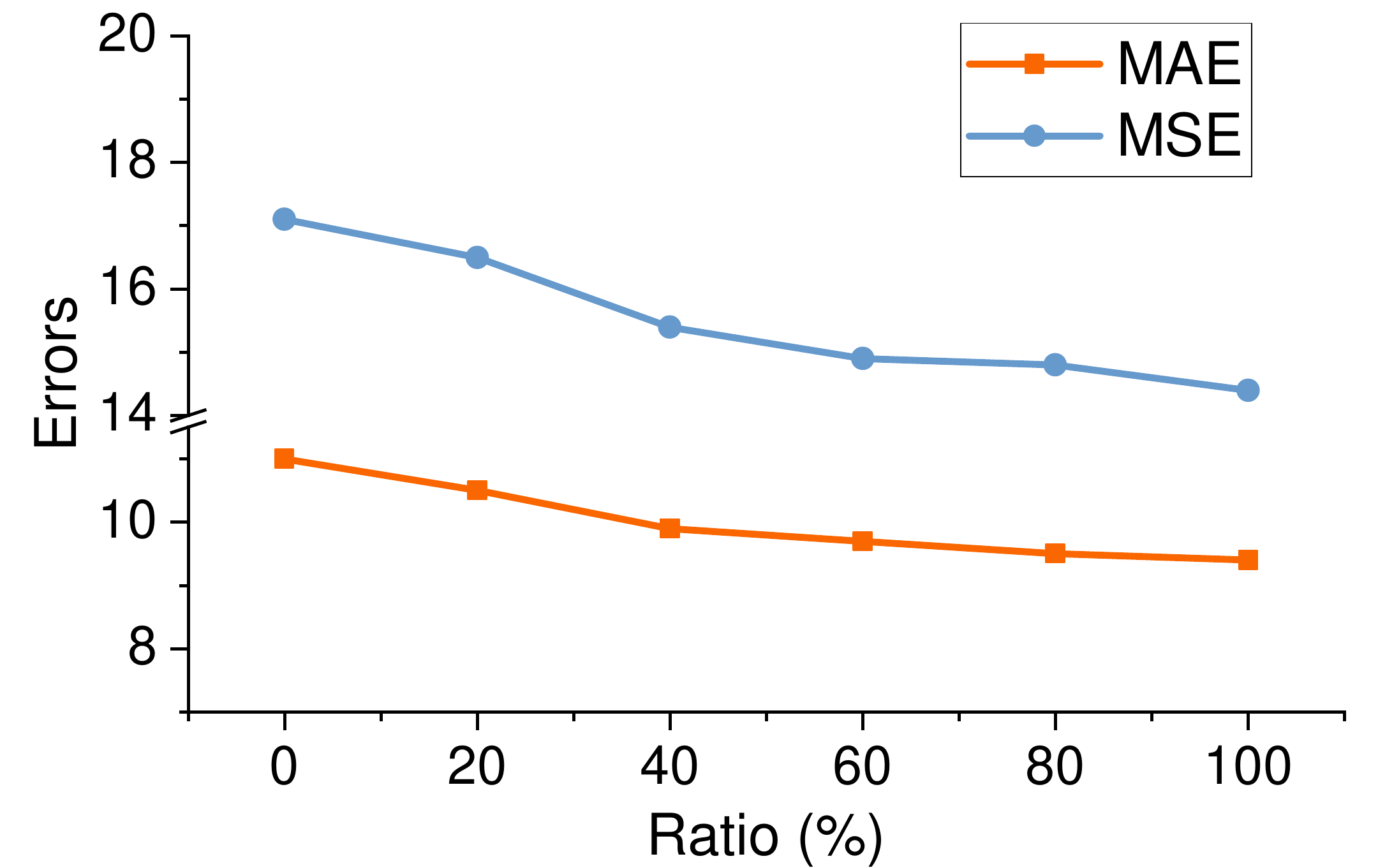}
	} 
	\caption{The fine-tuning SFCN's results on UCF-QNRF and SHT B datasets by pre-training different pre-trained GCC data volumes. ``0\%'' means the model do not use any GCC data to pre-train, namely Pre-ImgNt.} 
	\label{Fig-curve-gcc} 
\end{figure}

For the first factor, more images mean that the pre-trained data is more diverse. Thus, we implement Pre-GCC SFCN models using $20\%$, $40\%$, $60\%$, $80\%$ and $100\%$ GCC data, respectively. Fig. \mbox{\ref{Fig-curve-gcc}} demonstrates the fine-tuning SFCN's estimation errors on UCF-QNRF and SHT B datasets by using different pre-trained GCC data volumes. From the figure, as the pre-training data gradually increases, the estimation errors (MAE and MSE) become smaller. It means that more diverse data can better help fine-tune the model on real-world data.

For the second factor, different combinations of scene levels indicate different density distributions in the pre-trained data. In GCC, all scenes are divided into nine categories, denoting $L0$, $L1$, ..., and $L8$. Here, we merge them into four categories: $\{L0, L1, L2\}$, $\{L3, L4\}$, $\{L5, L6\}$ and $\{L7, L8\}$. To eliminate the impact of data volume, we sample $1,800$ images from the 4 classes as the pre-trained data. Table \mbox{\ref{Table-pre-sl}} reports the performance (MAE/MSE) of the fine-tuning SFCN's results on UCF-QNRF and SHT B datasets. Since UCF-QNRF is an extremely congested dataset, the results of pre-training on $\{L5, L6\}$ and $\{L7, L8\}$ are better than that of pre-training on sparse crowd scenes (namely $\{L0, L1, L2\}$ and $\{L3, L4\}$. Similarly, SHT B's density range is in $[9,578]$ (reported in Table \mbox{\ref{Table-compare}}), so when pre-training on $\{L5, L6\}$ (the density range is in $[0,600]$ and $[0,1000]$), the errors are the lowest (MAE/MSE of 9.1/14.8). According to the above results, we find that when the density distribution of pre-trained data is closer to that of real data, the fine-tuning performance will be better.

\begin{table}[htbp]
	
	\centering
	\caption{The fine-tuning SFCN's results (MAE/MSE) on UCF-QNRF and SHT B datasets by pre-training GCC data with different scene levels.}
	
	\begin{tabular}{c|c|c}
		\whline
		Pre-trained data &QNRF &SHT B\\
		\whline
		$\{L0, L1, L2\}$  &131.4/223.5  &10.8/17.4 \\
		\hline
		$\{L3, L4\}$   &121.6/201.9 & 9.3/15.0 \\
		\hline
		$\{L5, L6\}$   &117.0/\textbf{196.1} & \textbf{9.1}/\textbf{14.8} \\
		\hline
		$\{L7, L8\}$   &\textbf{115.0}/207.6 & 9.8/17.1 \\
		\whline	
	\end{tabular}
	\label{Table-pre-sl}
\end{table}

Finally, we explore the effect of GCC data's luminance changes on the fine-tuning results. To be specific, GCC is roughly divided into high-luminance and low-luminance data according to the time of shooting (high-luminance range is 6:00 $\sim$ 17:59 and the others are low-luminance data). To eliminate the impact of data volume, we sample $2,500$ images from the two types of data classes as the pre-trained data. Table \mbox{\ref{Table-pre-l}} shows the fine-tuning SFCN's results (MAE/MSE) on UCF-QNRF and SHT B datasets. The final estimation errors using low-luminance data is larger than that of using high-luminance data. The main reason is that the real-world datasets (UCF-QNRF and SHT B) rarely contain low-luminance data. In other words, the high-luminance GCC data is closer to the real data.

\begin{table}[htbp]
	
	\centering
	\caption{The fine-tuning SFCN's results (MAE/MSE) on UCF-QNRF and SHT B datasets by pre-training high-/low- luminance GCC data.}
	
	\begin{tabular}{c|c|c}
		\whline
		Pre-trained data &QNRF &SHT B\\
		\whline
		Low luminance  &126.7/231.9  &10.6/16.1 \\
		\hline
		High luminance   &\textbf{117.2}/\textbf{199.4} & \textbf{9.2}/\textbf{14.3} \\
		\whline	
	\end{tabular}
	\label{Table-pre-l}
\end{table}

In summary, to further prompt the performance of Pre-GCC, we may need as much pre-training data as possible that is more similar to the real data. It will be an interesting question about how to select the proper pre-training data.

\subsubsection{Comparison with the SOTA Methods}

\begin{table*}[htbp]
	\centering
	\small
	\caption{The comparison with the state-of-the-art performance on the five real datasets.}
	\setlength{\tabcolsep}{1.41mm}{
		\begin{tabular}{cIc|cIc|cIc|cIc|cIc|c|c|c|c|c}
			\whline
			\multirow{2}{*}{Method}	 &\multicolumn{2}{cI}{UCF-QNRF} &\multicolumn{2}{cI}{SHT A} &\multicolumn{2}{cI}{SHT B}  &\multicolumn{2}{cI}{UCF\_CC\_50} & \multicolumn{6}{c}{WorldExpo'10 (MAE)}\\
			\cline{2-15} 
			& MAE &MSE & MAE &MSE & MAE &MSE  & MAE &MSE  &S1 &S2 &S3 &S4 &S5 &Avg.\\
			\whline
			MCNN~\citep{zhang2016single}  &277 &426 &110.2 &173.2 &26.4 &41.3  &377.6 &509.1 &3.4&20.6& 12.9	& 13.0& 8.1& 11.6\\
			\hline
			Switching-CNN~\citep{sam2017switching} &- &- &90.4 &135.0 &21.6  &33.6 &318.1 &439.2  &4.4&15.7&10.0&11.0&5.9&9.4\\
			\hline
			CP-CNN~\citep{sindagi2017generating} &- &- &73.6 &106.4 &20.1&30.1 &298.8 &320.9 &2.9& 14.7& 10.5& 10.4& 5.8& 8.86 \\
			\hline
			ACSCP~\citep{shen2018crowd}  &- &- &75.7 &102.7 &17.2 &27.4 &291.0 &404.6 &2.8& 14.05& 9.6&\textbf{8.1}&\textbf{2.9}& \textbf{7.5} \\
			\hline
			CSRNet~\citep{li2018csrnet}   &- &- &68.2 &115.0 &10.6 &16.0 &266.1 &397.5  &2.9&\textbf{11.5}&\textbf{8.6}& 16.6& 3.4& 8.6 \\
			\hline
			DRSAN~\citep{liu2018crowd} &- &- &69.3 &\textbf{96.4} &11.1 &18.2 &219.2 &\textbf{250.2} &  2.6&	11.8& 10.3& 10.4& 3.7& 7.76\\
			\hline			
			SANet~\citep{cao2018scale} &- &- &67.0 &104.5 &17.0 &8.4&258.4 &334.9  &2.6& 13.2&9.0 &13.3& 3.0& 8.2\\
			\hline
			CL~\citep{idrees2018composition} &132 &191 &- &- &-&- &- &- &-&-&-&-&-&- \\
			\hline		
			ic-CNN~\citep{ranjan2018iterative} &- &- &68.5 &116.2 &10.7&16.0 &260.9 &365.5 &17.0&	12.3&9.2&\textbf{8.1}&4.7&10.3 \\
			\hline								
			SFCN\dag\, with Pre-GCC &\textbf{102.0} &\textbf{171.4}&\textbf{64.8} &107.5 &\textbf{7.6}  &\textbf{13.0} &\textbf{214.2} &318.2 &\textbf{1.8} &17.5&11.1 &13.5 &3.0 &9.4  \\
			\whline			
		\end{tabular}
	}\label{Table-sota}
\end{table*}

For comparison with other State-of-the-art methods, we conduct the experiments of SFCN\dag \, with the Pre-GCC strategy on five mainstream crowd counting datasets, namely UCF-QNRF, SHT A, SHT B, UCF\_CC\_50 and WorldExpo'10. Table \ref{Table-sota} reports the results of them. Our proposed method refreshes the six records in all nine metrics of the five datasets. To be specific, we achieve the best MAE performance on UCF-QNRF (\textbf{102.0}), SHT A (\textbf{64.8}), SHT B (\textbf{7.6}), UCF\_CC\_50 (\textbf{214.2}).

\subsection{Results of Domain-adaptation Crowd Understanding}

In this section, we conduct the adaptation experiments and further analyze the effectiveness of the proposed CycleGAN-based methods. 

\subsubsection{Implementation Details}

\label{para_da}

Like Section \mbox{\ref{para_sc}}, we randomly select 10\% training data of the real domain as the validation set to find the best model. During the training phase, $\alpha$, $\beta$ and $\lambda$ in Eq. \mbox{\ref{all_loss}} are set as $1$, $0.1$ and $0.01$, respectively.  SE CycleGAN's and SFCN's training parameter is same as the original CycleGAN and Section \mbox{\ref{para_sc}}, respectively. $D_{dc}$'s learning rate is set as $10^{-4}$. 

In Section \mbox{\ref{ad_SFCN}}, we introduce Scene Regularization (SR) to select the proper images to avoid negative adaptation. Here, Table \mbox{\ref{Table-filter}} shows the concrete filter condition for adaptation to the five real datasets. Specifically, ratio range means that the numbers of people in selected images should be in a specific range. For example, during adaptation to SHT A, there is a candidate image with level 0$\sim$4000, containing 800 people. According to the ratio range of 0.5$\sim$1, since 800 is not in 2000$\sim$4000 (namely 0.5*4000 $\sim$1*4000), the image can not be selected. In other words, the ratio range is a restriction in terms of congestion. 

\begin{table}[htbp]
	\scriptsize
	\centering
	\caption{Filter condition on eight real datasets.}
	\setlength{\tabcolsep}{1.6mm}{
		\begin{tabular}{c|c|c|c|c|c}
			\whline
			Target Dataset  & level & time & weather & count range & ratio range\\
			\whline
			SHT A  &4,5,6,7,8 & 6:00$\sim$19:59 & 0,1,3,5,6 &25$\sim$4000 & 0.5$\sim$1 \\
			\hline
			SHT B  &1,2,3,4,5 & 6:00$\sim$19:59 & 0,1,5,6 &10$\sim$600 & 0.3$\sim$1 \\
			\hline
			UCF\_CC\_50  &5,6,7,8 & 8:00$\sim$17:59 & 0,1,5,6 &400$\sim$4000 & 0.6$\sim$1\\
			\hline
			UCF-QNRF  &4,5,6,7,8 & 5:00$\sim$20:59 & 0,1,5,6 &400$\sim$4000 & 0.6$\sim$1\\
			\hline
			WorldExpo'10  &2,3,4,5,6 & 6:00$\sim$18:59 & 0,1,5,6 &0$\sim$1000 & 0$\sim$1\\
			\whline
		\end{tabular}
	}
	\label{Table-filter}
\end{table}

Other explanations of Arabic numerals in the table is listed as follows:

[Level Categories] 0: 0$\sim$10, 1: 0$\sim$25, 2: 0$\sim$50, 3: 0$\sim$100, 4: 0$\sim$300, 5: 0$\sim$600, 6: 0$\sim$1k, 7: 0$\sim$2k and 8: 0$\sim$4k. 

[Weather Categories] 0: clear, 1: clouds, 2: rain, 3: foggy, 4: thunder, 5: overcast and 6: extra sunny. 


\subsubsection{Adaptation Performance on Real-world Datasets}

In this section, we conduct the adaptation experiments from GCC dataset to five mainstream real-world counting datasets: ShanghaiTech A/B \citep{zhang2016single}, UCF\_CC\_50 \citep{idrees2013multi}, UCF-QNRF \citep{idrees2018composition}, WorldExpo'10 \citep{zhang2016data} and a real-word crowd segmentation dataset, CityScapes \mbox{\citep{cordts2016cityscapes}}. For the best performance, all models adopt the Scene/Density Regularization mentioned in Section \ref{SDR}. Notably, each model is explained as follows:

\textbf{NoAdpt:} Train SFCN on the original GCC and evaluate on the real dataset.

\textbf{CycleGAN:} Translate GCC images to photo-realistic data using CycleGAN, and then train SFCN on them.

\textbf{SE CycleGAN:} Translate GCC images to photo-realistic data using SE CycleGAN, and then train SFCN on them. It is the method of the conference version \mbox{\citep{wang2019learning}}.

\textbf{SE CycleGAN (Joint Training, JT):} Jointly train SE CycleGAN (introducing feature-level adversarial learning) model and SFCN.

\begin{table*}[htbp]
	\centering
	\small
	\caption{The counting performance of no adaptation (No Adpt), CycleGAN \citep{zhu2017unpaired}, SE CycleGAN and SE CycleGAN (Joint Training, JT) on the five real-world datasets.}
	\setlength{\tabcolsep}{2.1mm}{
		\begin{tabular}{c|cIc|c|c|cIc|c|c|cIc|c|c|c}
			\hline
			\multirow{2}{*}{Method}	&\multirow{2}{*}{DA} &\multicolumn{4}{cI}{SHT A} &\multicolumn{4}{cI}{SHT B} &\multicolumn{4}{c}{UCF\_CC\_50}\\
			\cline{3-14} 
			& & MAE &MSE &PSNR &SSIM &MAE & MSE &PSNR &SSIM  &MAE & MSE &PSNR &SSIM \\
			\hline
			NoAdpt  &\xmark &160.0 &216.5 &19.01 &0.359 &22.8 &30.6 &24.66 &0.715 &487.2 &689.0 &17.27 &0.386 \\
			\hline
			CycleGAN &\rmark &143.3 &204.3 &\textbf{19.27} &0.379  &25.4 &39.7 &24.60 &0.763  &404.6 &548.2 &\textbf{17.34} &0.468 \\
			\hline	
			SE CycleGAN  &\rmark&123.4 &193.4 &18.61 &0.407 &19.9 &28.3 &24.78 &0.765 &373.4 &528.8 &17.01 &\textbf{0.743}  \\
			\hline	
			SE CycleGAN (JT)  &\rmark&\textbf{119.6} &\textbf{189.1} &18.69 &\textbf{0.429} &\textbf{16.4} &\textbf{25.8} &\textbf{26.17} &\textbf{0.786} &\textbf{370.2} &\textbf{512.0} &17.11 &0.689  \\			
			\hline
		\end{tabular}
		\vspace{0.2cm}
	}
	
	\setlength{\tabcolsep}{2.2mm}{
		\begin{tabular}{c|cIc|c|c|cIc|c|c|c|c|c}
			\hline
			\multirow{2}{*}{Method}	&\multirow{2}{*}{DA} &\multicolumn{4}{cI}{UCF-QNRF} &\multicolumn{6}{c}{WorldExpo'10 (MAE)}\\
			\cline{3-12} 
			& & MAE &MSE &PSNR &SSIM &S1 &S2 &S3 &S4 &S5 &Avg.  \\
			\hline
			NoAdpt  &\xmark &275.5 &458.5 &20.12 &0.554 &4.4 &87.2 &59.1  &51.8 &11.7 &42.8 \\
			\hline
			CycleGAN &\rmark &257.3 &400.6 &20.80 &0.480  &4.4 &69.6 &49.9 &29.2  &9.0 &32.4   \\
			\hline	
			SE CycleGAN  &\rmark&230.4 &\textbf{384.5} &21.03 &\textbf{0.660} &4.3 &59.1 &43.7  &\textbf{17.0} &7.6 &26.3  \\
			\hline
			SE CycleGAN (JT)  &\rmark&\textbf{225.9} &385.7 &\textbf{21.10} &0.642 &\textbf{4.2} &\textbf{49.6} &\textbf{41.3}  &19.8 &\textbf{7.2} &\textbf{24.4}  \\
			\hline
		\end{tabular}
	}
	\label{Table-DA}
\end{table*}

\begin{figure*}
	\centering
	\includegraphics[width=0.98\textwidth]{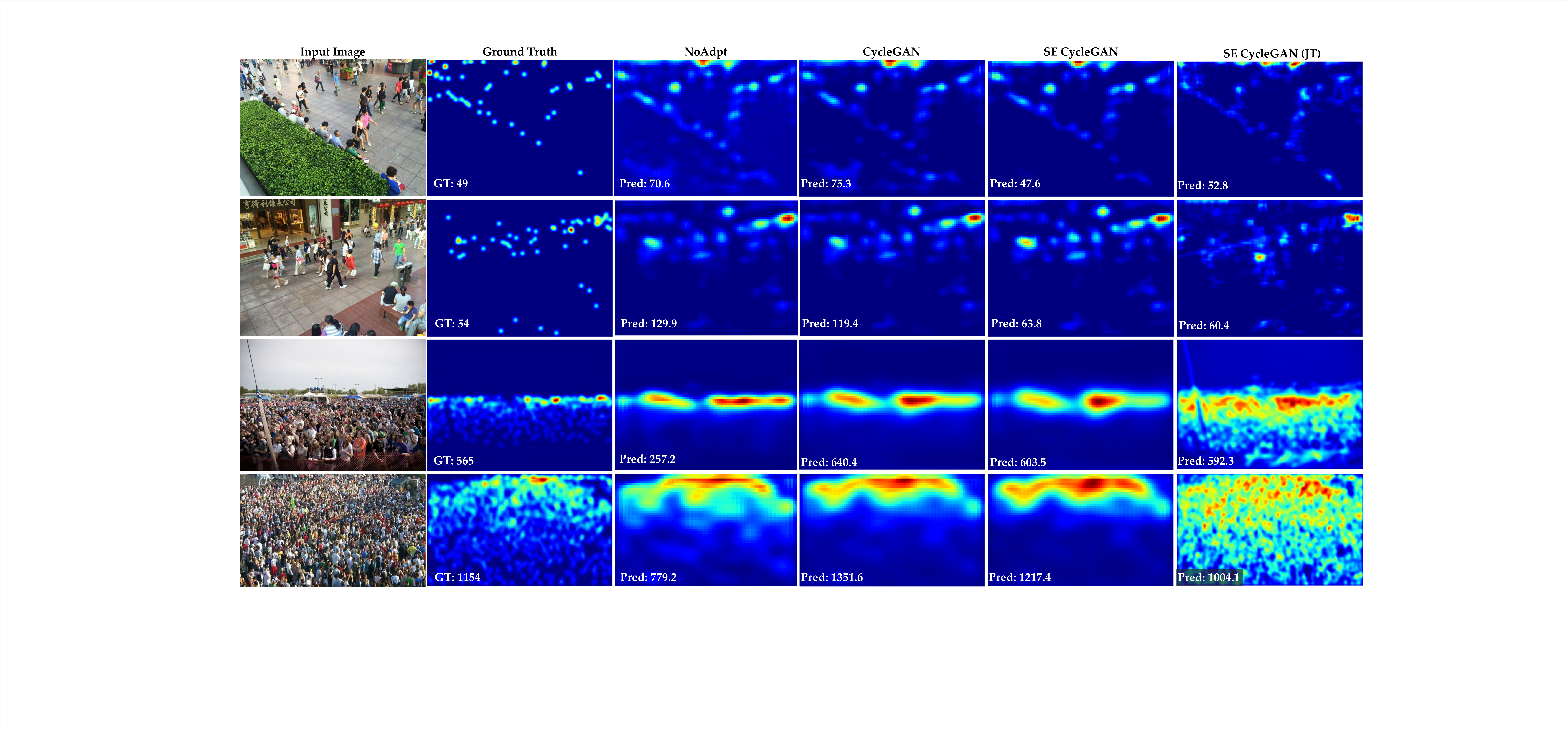}
	\caption{The demonstration of different methods on SHT dataset. ``GT'' and ``Pred'' represent the labeled and predicted count, respectively. }\label{Fig-dare}
\end{figure*}

\vspace{0.2cm}
\noindent\textbf{Crowd Counting via Domain Adaptation \,\,}

Table \mbox{\ref{Table-DA}} shows the four metrics (only MAE on WorldExpo'10) of the No Adaptation (NoAdpt), CycleGAN, the proposed SE CycleGAN and SE CycleGAN (Joint Training, JT). From it, we find the results after adaptation are far better than that of no adaptation, which indicates the adaptation can effectively reduce the domain gaps between synthetic and real-world data. After embedding SSIM loss in CycleGAN, almost all performances are improved on five datasets. There are only two reductions of PSNR on Shanghai Tech A and UCF\_CC\_50. In general, the proposed SE CycleGAN outperforms the original CycleGAN. When utilizing jointing training, the performance is increased in most metrics, which means that adversarial learning and joint training can further reduce the domain gap between translated synthetic images and real-world data.

In addition, we find the counting results on SHT B achieves a good level (MAE/MSE of 16.4/25.8), even outperform some early supervised methods \citep{zhang2016single, sindagi2017cnn, sam2017switching, sindagi2017generating,liu2018decidenet}. The main reasons are: 1) the real data is strongly consistent, which is captured by the same sensors; 2) the data has high image clarity. The two characteristics guarantee that the SE CycleGAN's adaptation on SHT B is more effective than others. 

Fig. \ref{Fig-dare} demonstrates four groups of visualized results on SHT A and B dataset. Compared with NoAdpt, the map quality via CycleGAN has a significant improvement. Row 1 and 2 demonstrate the Part B visualization results. We find the predicted maps are very close to the groundtruth. When jointly training SE CycleGAN and SFCN, we find the misestimation in the background region can be effectively alleviated compared with the original SE CycleGAN. However, for the extremely congested scenes (in Row 3 and 4), the predicted maps are very far from the ground truth. We think the main reason is that the translated images lose the details (such as texture, sharpness and edge) in high-density regions. 

\begin{figure*}
	\centering
	\includegraphics[width=0.98\textwidth]{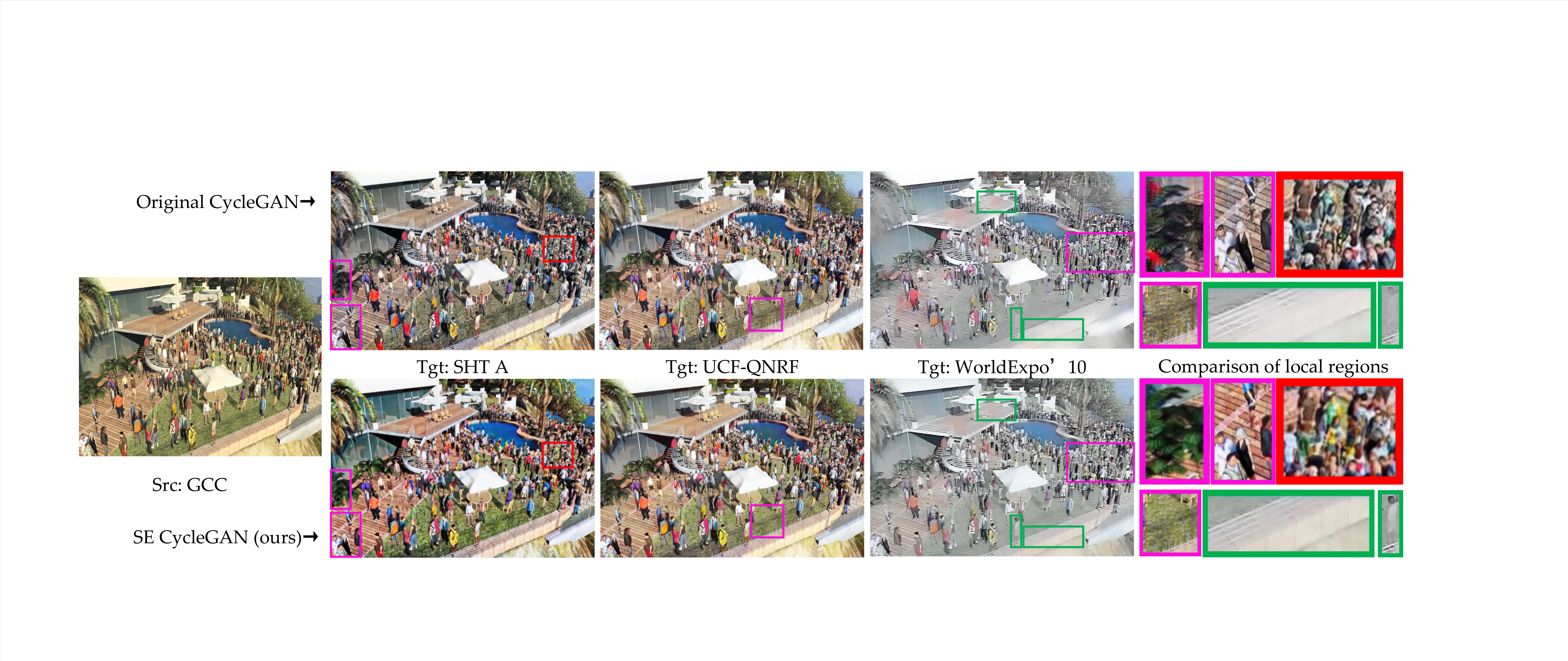}
	\caption{The visualization comparisons of CycleGAN and SE CycleGAN. }\label{Fig-com}
\end{figure*}

\vspace{0.2cm}
\noindent\textbf{Crowd Segmentation via Domain Adaptation \,\,}
	
Considering that there is no real-world dataset for crowd segmentation, we transform a semantic segmentation dataset (CityScapes, \mbox{\cite{cordts2016cityscapes}}) to a crowd segmentation dataset. To be specific, the ``pedestrian'' class to generate the crowd mask and the other objects are treated as ``background''. Same as the counting task, we conduct two groups of experiments: NoAdpt and SE CycleGAN (Joint training, JT). The results are reported in Table \mbox{\ref{Table-seg}}. From the table, after introducing domain adaptation, mIoU is increased by 10.1\% compared with NoAdapt, which evidences that the proposed framework is also suitable for the domain-adaptation-style crowd segmentation task. 

\begin{table}[htbp]
	
	\centering
	\caption{The segmentation performance (\%) of no adaptation (NoAdpt) and SE CycleGAN (Joint Training, JT) on CityScapes dataset.}
	\begin{tabular}{c|c|c|c}
		\whline
		Method	&fg & bg & mIoU\\
		\whline
		NoAdpt &98.9 &20.7& 59.8   \\
		\hline
		SE CycleGAN (JT)  &96.3 &35.4 &65.9  \\
		\whline
	\end{tabular}
	\label{Table-seg}
\end{table}

\subsubsection{The Effectiveness of SDR}
Here, we compare the performance of three models (No Adpt, CycleGAN and SE CycleGAN) without Scene/Density Regularization (SDR) and with SDR. Table \ref{Table-dsr} reports the performance of with or without SDR on SHT A dataset. From the results in the first column, we find these two adaptation methods cause some side effects. In fact, they do not produce ideal translated images. When introducing SDR, the nonexistent synthetic scenes in the real datasets are filtered out, which improves the domain adaptation performance.

\begin{table}[htbp]
	
	\centering
	\caption{The results of NoAdpt, CycleGAN \citep{zhu2017unpaired} and SE CycleGAN on SHT A.}
	\begin{tabular}{c|c|c}
		\whline
		Method	&w/o SDR &with SDR\\
		\whline
		NoAdpt &\textbf{163.6}/244.5 &\underline{160.0}/\underline{216.5}   \\
		\hline
		CycleGAN  &180.1/290.3 &\underline{143.3}/\underline{204.3}  \\
		\hline
		SE CycleGAN  &169.8/\textbf{230.2} &\underline{\textbf{123.4}}/\underline{\textbf{193.4}}  \\
		\whline
	\end{tabular}
	\label{Table-dsr}
\end{table}

\subsubsection{Analysis of SSIM Embedding}

SSIM Embedding can guarantee the original synthetic and reconstructed images have high structural similarity (SS), which prompts two generators' translation for images to maintain a certain degree of SS during the training process. Fig. \ref{Fig-com}, demonstrate the translated images from GCC to the three real-world datasets. ``Src'' and ``Tgt'' represent the source domain (synthetic data) and the target domain (real-world data). The top row shows the results of the original CycleGAN and the bottom is the results of the proposed SE CycleGAN.

We compare some obvious differences between CycleGAN and SE CycleGAN (ours) and mark them up with rectangular boxes. To be specific, ours can produce a more consistent image than the original CycleGAN in the green boxes. As for the red boxes, CycleGAN loses more texture features than ours. For the purple boxes, we find that CycleGAN produces some abnormal color values, but SE CycleGAN performs better than it. For the regions covered by blue boxes, SE CycleGAN maintains the contrast of the original image than CycleGAN in a even better fashion. 

In general, from the visualization results, the proposed SE CycleGAN generates more high-quality crowd scenes than the original CycleGAN. The complete translation results are available at this website \footnote{http://share.crowdbenchmark.com:2443/home/Translation\_Results}.


\section{Conclusion and Outlook}
\label{conclusion}
In this paper, we focus on promoting the performance of crowd understanding in the wild via utilizing the synthetic data. Exploiting the generated data, we then propose two effective ways (pre-training scheme and domain adaptation) to improve the counting performance in the real world significantly. The proposed pre-training scheme provides a better installation parameter than the traditional strategy, namely pre-training on ImageNet. Experiments show that the counting performance is improved by an average of 12\%. The presented domain adaptation provides a new direction for crowd understanding, which liberates humans from the tedious labeling work. By the joint training for adaptation and source-domain crowd understanding, the trained crowd model works better than traditional no adaptation method in the real-world data. To be specific, in some typical subservience scenes (ShanghaiTech Part B, WorldExpo'10, UCSD and Mall), the estimations of domain adaptation are very close to that of the traditional supervised learning. 

According to the results of experiments from this paper, we think there are some interesting directions in the crowd understanding:
\begin{enumerate}
	\item[1)] \textbf{Data generation \,\,} Based on the open-sourced tools, the researchers are allowed to re-develop customized software for generating synthetic image or video data including but not limited to the following tasks: object counting/localization/tracking (crowd, vehicle, \emph{etc.}), crowd instance segmentation, crowd flow analysis, group detection, person re-identification, anomaly event detection, and human trajectory prediction. 
	\item[2)] \textbf{Domain-adaptive crowd understanding \,\,} The traditional supervised learning requires a large amount of labeled data, which hinders the landing of the crowd model in the real world. Considering this problem, we think domain-adaptive crowd counting is a more practical research area than supervised learning: unsupervised/few-shot domain adaptation will reduce the cost of collecting and annotating real scene data.
\end{enumerate}

In future work, we will focus on the two types as mentioned above of tasks and attempt to promote the practical application of crowd understanding in the real world.


\bibliographystyle{spbasic}      
\bibliography{ref}   

\begin{thebibliography}{78}
\providecommand{\natexlab}[1]{#1}
\providecommand{\url}[1]{{#1}}
\providecommand{\urlprefix}{URL }
\expandafter\ifx\csname urlstyle\endcsname\relax
  \providecommand{\doi}[1]{DOI~\discretionary{}{}{}#1}\else
  \providecommand{\doi}{DOI~\discretionary{}{}{}\begingroup
  \urlstyle{rm}\Url}\fi
\providecommand{\eprint}[2][]{\url{#2}}

\bibitem[{Abu-El-Haija et~al.(2016)Abu-El-Haija, Kothari, Lee, Natsev,
  Toderici, Varadarajan, and Vijayanarasimhan}]{abu2016youtube}
Abu-El-Haija S, Kothari N, Lee J, Natsev P, Toderici G, Varadarajan B,
  Vijayanarasimhan S (2016) Youtube-8m: A large-scale video classification
  benchmark. arXiv preprint arXiv:160908675

\bibitem[{Babu~Sam et~al.(2018)Babu~Sam, Sajjan, Venkatesh~Babu, and
  Srinivasan}]{babu2018divide}
Babu~Sam D, Sajjan NN, Venkatesh~Babu R, Srinivasan M (2018) Divide and grow:
  Capturing huge diversity in crowd images with incrementally growing cnn. In:
  Proceedings of the IEEE Conference on Computer Vision and Pattern
  Recognition, pp 3618--3626

\bibitem[{Bak et~al.(2018)Bak, Carr, and Lalonde}]{bak2018domain}
Bak S, Carr P, Lalonde JF (2018) Domain adaptation through synthesis for
  unsupervised person re-identification. arXiv preprint arXiv:180410094

\bibitem[{Cao et~al.(2018)Cao, Wang, Zhao, and Su}]{cao2018scale}
Cao X, Wang Z, Zhao Y, Su F (2018) Scale aggregation network for accurate and
  efficient crowd counting. In: Proceedings of the European Conference on
  Computer Vision, pp 734--750

\bibitem[{Chan and Vasconcelos(2009)}]{chan2009bayesian}
Chan AB, Vasconcelos N (2009) Bayesian poisson regression for crowd counting.
  In: 2009 IEEE 12th international conference on computer vision, IEEE, pp
  545--551

\bibitem[{Chan et~al.(2008)Chan, Liang, and Vasconcelos}]{chan2008privacy}
Chan AB, Liang ZSJ, Vasconcelos N (2008) Privacy preserving crowd monitoring:
  Counting people without people models or tracking. In: Proceedings of the
  IEEE conference on Computer Vision and Pattern Recognition, pp 1--7

\bibitem[{Chan et~al.(2009)Chan, Morrow, Vasconcelos et~al.}]{chan2009analysis}
Chan AB, Morrow M, Vasconcelos N, et~al. (2009) Analysis of crowded scenes
  using holistic properties. In: Performance Evaluation of Tracking and
  Surveillance workshop at CVPR, pp 101--108

\bibitem[{Chen et~al.(2012)Chen, Loy, Gong, and Xiang}]{chen2012feature}
Chen K, Loy CC, Gong S, Xiang T (2012) Feature mining for localised crowd
  counting. In: Proceedings of the British Machine Vision Conference, vol~1,
  p~3

\bibitem[{Cordts et~al.(2016)Cordts, Omran, Ramos, Rehfeld, Enzweiler,
  Benenson, Franke, Roth, and Schiele}]{cordts2016cityscapes}
Cordts M, Omran M, Ramos S, Rehfeld T, Enzweiler M, Benenson R, Franke U, Roth
  S, Schiele B (2016) The cityscapes dataset for semantic urban scene
  understanding. In: Proceedings of the IEEE conference on computer vision and
  pattern recognition, pp 3213--3223

\bibitem[{Deng et~al.(2009)Deng, Dong, Socher, Li, Li, and
  Fei-Fei}]{deng2009imagenet}
Deng J, Dong W, Socher R, Li LJ, Li K, Fei-Fei L (2009) Imagenet: A large-scale
  hierarchical image database. In: Proceedings of the IEEE conference on
  Computer Vision and Pattern Recognition, pp 248--255

\bibitem[{Dong et~al.(2007)Dong, Parameswaran, Ramesh, and
  Zoghlami}]{dong2007fast}
Dong L, Parameswaran V, Ramesh V, Zoghlami I (2007) Fast crowd segmentation
  using shape indexing. In: 2007 IEEE 11th International Conference on Computer
  Vision, IEEE, pp 1--8

\bibitem[{Dosovitskiy et~al.(2017)Dosovitskiy, Ros, Codevilla, Lopez, and
  Koltun}]{Dosovitskiy17}
Dosovitskiy A, Ros G, Codevilla F, Lopez A, Koltun V (2017) {CARLA}: {An} open
  urban driving simulator. In: Proceedings of the 1st Annual Conference on
  Robot Learning, pp 1--16

\bibitem[{Everingham et~al.(2015)Everingham, Eslami, Van~Gool, Williams, Winn,
  and Zisserman}]{everingham2015pascal}
Everingham M, Eslami SA, Van~Gool L, Williams CK, Winn J, Zisserman A (2015)
  The pascal visual object classes challenge: A retrospective. International
  journal of computer vision 111(1):98--136

\bibitem[{Fu et~al.(2015)Fu, Xu, Li, Liu, Ye, and Zhu}]{fu2015fast}
Fu M, Xu P, Li X, Liu Q, Ye M, Zhu C (2015) Fast crowd density estimation with
  convolutional neural networks. Engineering Applications of Artificial
  Intelligence 43:81--88

\bibitem[{Gao et~al.(2019)Gao, Lin, Zhao, Wang, Gao, and Wen}]{gao2019c}
Gao J, Lin W, Zhao B, Wang D, Gao C, Wen J (2019) C$^3$ framework: An
  open-source pytorch code for crowd counting. arXiv preprint arXiv:190702724

\bibitem[{{Gao} et~al.(2019){Gao}, {Wang}, and {Li}}]{8723079}
{Gao} J, {Wang} Q, {Li} X (2019) Pcc net: Perspective crowd counting via
  spatial convolutional network. IEEE Transactions on Circuits and Systems for
  Video Technology pp 1--1, \doi{10.1109/TCSVT.2019.2919139}

\bibitem[{Goodfellow et~al.(2014)Goodfellow, Pouget-Abadie, Mirza, Xu,
  Warde-Farley, Ozair, Courville, and Bengio}]{goodfellow2014generative}
Goodfellow I, Pouget-Abadie J, Mirza M, Xu B, Warde-Farley D, Ozair S,
  Courville A, Bengio Y (2014) Generative adversarial nets. In: Proceedings of
  the Advances in Neural Information Processing Systems, pp 2672--2680

\bibitem[{He et~al.(2016)He, Zhang, Ren, and Sun}]{he2016deep}
He K, Zhang X, Ren S, Sun J (2016) Deep residual learning for image
  recognition. In: Proceedings of the IEEE conference on Computer Vision and
  Pattern Recognition, pp 770--778

\bibitem[{Huang et~al.(2017)Huang, Liu, Van Der~Maaten, and
  Weinberger}]{huang2017densely}
Huang G, Liu Z, Van Der~Maaten L, Weinberger KQ (2017) Densely connected
  convolutional networks. In: Proceedings of the IEEE conference on Computer
  Vision and Pattern Recognition, pp 4700--4708

\bibitem[{Idrees et~al.(2013)Idrees, Saleemi, Seibert, and
  Shah}]{idrees2013multi}
Idrees H, Saleemi I, Seibert C, Shah M (2013) Multi-source multi-scale counting
  in extremely dense crowd images. In: Proceedings of the IEEE conference on
  Computer Vision and Pattern Recognition, pp 2547--2554

\bibitem[{Idrees et~al.(2018)Idrees, Tayyab, Athrey, Zhang, Al-Maadeed,
  Rajpoot, and Shah}]{idrees2018composition}
Idrees H, Tayyab M, Athrey K, Zhang D, Al-Maadeed S, Rajpoot N, Shah M (2018)
  Composition loss for counting, density map estimation and localization in
  dense crowds. arXiv preprint arXiv:180801050

\bibitem[{Jia et~al.(2014)Jia, Shelhamer, Donahue, Karayev, Long, Girshick,
  Guadarrama, and Darrell}]{jia2014caffe}
Jia Y, Shelhamer E, Donahue J, Karayev S, Long J, Girshick R, Guadarrama S,
  Darrell T (2014) Caffe: Convolutional architecture for fast feature
  embedding. In: Proceedings of the 22nd ACM international conference on
  Multimedia, ACM, pp 675--678

\bibitem[{Jiang et~al.(2019)Jiang, Xiao, Zhang, Zhen, Cao, Doermann, and
  Shao}]{Jiang_2019_CVPR}
Jiang X, Xiao Z, Zhang B, Zhen X, Cao X, Doermann D, Shao L (2019) Crowd
  counting and density estimation by trellis encoder-decoder networks. In: The
  IEEE Conference on Computer Vision and Pattern Recognition (CVPR), pp
  6133--6142

\bibitem[{Johnson-Roberson et~al.(2017)Johnson-Roberson, Barto, Mehta, Sridhar,
  Rosaen, and Vasudevan}]{Johnson-Roberson:2017aa}
Johnson-Roberson M, Barto C, Mehta R, Sridhar SN, Rosaen K, Vasudevan R (2017)
  Driving in the matrix: Can virtual worlds replace human-generated annotations
  for real world tasks? In: Proceedings of the IEEE International Conference on
  Robotics and Automation, pp 1--8

\bibitem[{Kang and Wang(2014)}]{kang2014fully}
Kang K, Wang X (2014) Fully convolutional neural networks for crowd
  segmentation. arXiv preprint arXiv:14114464

\bibitem[{{Kempka} et~al.(2016){Kempka}, {Wydmuch}, {Runc}, {Toczek}, and
  {Jaśkowski}}]{7860433}
{Kempka} M, {Wydmuch} M, {Runc} G, {Toczek} J, {Jaśkowski} W (2016) Vizdoom: A
  doom-based ai research platform for visual reinforcement learning. In: 2016
  IEEE Conference on Computational Intelligence and Games (CIG), pp 1--8

\bibitem[{Kingma and Ba(2014)}]{kingma2014adam}
Kingma DP, Ba J (2014) Adam: A method for stochastic optimization. arXiv
  preprint arXiv:14126980

\bibitem[{Li et~al.(2018{\natexlab{a}})Li, Lin, Zuo, Tang, and
  Yang}]{li2018visual}
Li C, Lin L, Zuo W, Tang J, Yang MH (2018{\natexlab{a}}) Visual tracking via
  dynamic graph learning. IEEE transactions on pattern analysis and machine
  intelligence 41(11):2770--2782

\bibitem[{Li et~al.(2014)Li, Chang, Wang, Ni, Hong, and Yan}]{li2014crowded}
Li T, Chang H, Wang M, Ni B, Hong R, Yan S (2014) Crowded scene analysis: A
  survey. IEEE transactions on circuits and systems for video technology
  25(3):367--386

\bibitem[{Li et~al.(2013)Li, Mahadevan, and Vasconcelos}]{li2013anomaly}
Li W, Mahadevan V, Vasconcelos N (2013) Anomaly detection and localization in
  crowded scenes. IEEE transactions on pattern analysis and machine
  intelligence 36(1):18--32

\bibitem[{Li et~al.(2017)Li, Chen, Nie, and Wang}]{li2017multiview}
Li X, Chen M, Nie F, Wang Q (2017) A multiview-based parameter free framework
  for group detection. In: Thirty-First AAAI Conference on Artificial
  Intelligence

\bibitem[{Li et~al.(2018{\natexlab{b}})Li, Zhang, and Chen}]{li2018csrnet}
Li Y, Zhang X, Chen D (2018{\natexlab{b}}) Csrnet: Dilated convolutional neural
  networks for understanding the highly congested scenes. In: Proceedings of
  the IEEE Conference on Computer Vision and Pattern Recognition, pp 1091--1100

\bibitem[{Lian et~al.(2019)Lian, Li, Zheng, Luo, and Gao}]{Lian_2019_CVPR}
Lian D, Li J, Zheng J, Luo W, Gao S (2019) Density map regression guided
  detection network for rgb-d crowd counting and localization. In: The IEEE
  Conference on Computer Vision and Pattern Recognition (CVPR), pp 1821--1830

\bibitem[{Lin et~al.(2014)Lin, Maire, Belongie, Hays, Perona, Ramanan,
  Doll{\'a}r, and Zitnick}]{lin2014microsoft}
Lin TY, Maire M, Belongie S, Hays J, Perona P, Ramanan D, Doll{\'a}r P, Zitnick
  CL (2014) Microsoft coco: Common objects in context. In: European conference
  on computer vision, Springer, pp 740--755

\bibitem[{Liu et~al.(2018{\natexlab{a}})Liu, Gao, Meng, and
  Hauptmann}]{liu2018decidenet}
Liu J, Gao C, Meng D, Hauptmann AG (2018{\natexlab{a}}) Decidenet: Counting
  varying density crowds through attention guided detection and density
  estimation. In: Proceedings of the IEEE Conference on Computer Vision and
  Pattern Recognition, pp 5197--5206

\bibitem[{Liu et~al.(2018{\natexlab{b}})Liu, Wang, Li, Ouyang, and
  Lin}]{liu2018crowd}
Liu L, Wang H, Li G, Ouyang W, Lin L (2018{\natexlab{b}}) Crowd counting using
  deep recurrent spatial-aware network. arXiv preprint arXiv:180700601

\bibitem[{Liu et~al.(2019)Liu, Salzmann, and Fua}]{Liu_2019_CVPR}
Liu W, Salzmann M, Fua P (2019) Context-aware crowd counting. In: The IEEE
  Conference on Computer Vision and Pattern Recognition (CVPR), pp 5099--5108

\bibitem[{Liu et~al.(2018{\natexlab{c}})Liu, van~de Weijer, and
  Bagdanov}]{liu2018leveraging}
Liu X, van~de Weijer J, Bagdanov AD (2018{\natexlab{c}}) Leveraging unlabeled
  data for crowd counting by learning to rank. arXiv preprint arXiv:180303095

\bibitem[{Long et~al.(2015)Long, Shelhamer, and Darrell}]{long2015fully}
Long J, Shelhamer E, Darrell T (2015) Fully convolutional networks for semantic
  segmentation. In: Proceedings of the IEEE Conference on Computer Vision and
  Pattern Recognition, pp 3431--3440

\bibitem[{Mahadevan et~al.(2010)Mahadevan, Li, Bhalodia, and
  Vasconcelos}]{mahadevan2010anomaly}
Mahadevan V, Li W, Bhalodia V, Vasconcelos N (2010) Anomaly detection in
  crowded scenes. In: 2010 IEEE Computer Society Conference on Computer Vision
  and Pattern Recognition, IEEE, pp 1975--1981

\bibitem[{Marsden et~al.(2016)Marsden, McGuinness, Little, and
  O'Connor}]{marsden2016fully}
Marsden M, McGuinness K, Little S, O'Connor NE (2016) Fully convolutional crowd
  counting on highly congested scenes. arXiv preprint arXiv:161200220

\bibitem[{Mehran et~al.(2009)Mehran, Oyama, and Shah}]{mehran2009abnormal}
Mehran R, Oyama A, Shah M (2009) Abnormal crowd behavior detection using social
  force model. In: 2009 IEEE Conference on Computer Vision and Pattern
  Recognition, IEEE, pp 935--942

\bibitem[{Onororubio and Lopezsastre(2016)}]{onororubio2016towards}
Onororubio D, Lopezsastre RJ (2016) Towards perspective-free object counting
  with deep learning pp 615--629

\bibitem[{Pan et~al.(2017)Pan, Shi, Luo, Wang, and Tang}]{pan2017spatial}
Pan X, Shi J, Luo P, Wang X, Tang X (2017) Spatial as deep: Spatial cnn for
  traffic scene understanding. arXiv preprint arXiv:171206080

\bibitem[{Paszke et~al.(2017)Paszke, Gross, Chintala, and
  Chanan}]{paszke2017pytorch}
Paszke A, Gross S, Chintala S, Chanan G (2017) Pytorch: Tensors and dynamic
  neural networks in python with strong gpu acceleration

\bibitem[{Popoola and Wang(2012)}]{popoola2012video}
Popoola OP, Wang K (2012) Video-based abnormal human behavior recognition—a
  review. IEEE Transactions on Systems, Man, and Cybernetics, Part C
  (Applications and Reviews) 42(6):865--878

\bibitem[{Qiu et~al.(2017)Qiu, Zhong, Zhang, Qiao, Xiao, Kim, Wang, and
  Yuille}]{qiu2017unrealcv}
Qiu W, Zhong F, Zhang Y, Qiao S, Xiao Z, Kim TS, Wang Y, Yuille A (2017)
  Unrealcv: Virtual worlds for computer vision. ACM Multimedia Open Source
  Software Competition

\bibitem[{Ranjan et~al.(2018)Ranjan, Le, and Hoai}]{ranjan2018iterative}
Ranjan V, Le H, Hoai M (2018) Iterative crowd counting. arXiv preprint
  arXiv:180709959

\bibitem[{Richter et~al.(2016)Richter, Vineet, Roth, and
  Koltun}]{Richter_2016_ECCV}
Richter SR, Vineet V, Roth S, Koltun V (2016) Playing for data: {G}round truth
  from computer games. In: Proceedings of the European Conference on Computer
  Vision, pp 102--118

\bibitem[{Richter et~al.(2017)Richter, Hayder, and Koltun}]{richter2017playing}
Richter SR, Hayder Z, Koltun V (2017) Playing for benchmarks. In: Proceedings
  of the International conference on computer vision, vol~2

\bibitem[{Ros et~al.(2016)Ros, Sellart, Materzynska, Vazquez, and
  Lopez}]{ros2016synthia}
Ros G, Sellart L, Materzynska J, Vazquez D, Lopez AM (2016) The synthia
  dataset: A large collection of synthetic images for semantic segmentation of
  urban scenes. In: Proceedings of the IEEE conference on computer vision and
  pattern recognition, pp 3234--3243

\bibitem[{Sam et~al.(2017)Sam, Surya, and Babu}]{sam2017switching}
Sam DB, Surya S, Babu RV (2017) Switching convolutional neural network for
  crowd counting. In: Proceedings of the IEEE Conference on Computer Vision and
  Pattern Recognition, vol~1, p~6

\bibitem[{Sam et~al.(2018)Sam, Sajjan, and Babu}]{sam2018divide}
Sam DB, Sajjan NN, Babu RV (2018) Divide and grow: Capturing huge diversity in
  crowd images with incrementally growing cnn. arXiv: Computer Vision and
  Pattern Recognition

\bibitem[{Sam et~al.(2019)Sam, Sajjan, Maurya, and Babu}]{sam2019almost}
Sam DB, Sajjan NN, Maurya H, Babu RV (2019) Almost unsupervised learning for
  dense crowd counting. In: Proceedings of the Thirty-Third AAAI Conference on
  Artificial Intelligence, Honolulu, HI, USA, vol~27

\bibitem[{Shah et~al.(2018)Shah, Dey, Lovett, and Kapoor}]{shah2018airsim}
Shah S, Dey D, Lovett C, Kapoor A (2018) Airsim: High-fidelity visual and
  physical simulation for autonomous vehicles. In: Field and service robotics,
  Springer, pp 621--635

\bibitem[{Shen et~al.(2018)Shen, Xu, Ni, Wang, Hu, and Yang}]{shen2018crowd}
Shen Z, Xu Y, Ni B, Wang M, Hu J, Yang X (2018) Crowd counting via adversarial
  cross-scale consistency pursuit. In: Proceedings of the IEEE Conference on
  Computer Vision and Pattern Recognition, pp 5245--5254

\bibitem[{Shi et~al.(2018)Shi, Zhang, Liu, Cao, Ye, Cheng, and
  Zheng}]{shi2018crowd}
Shi Z, Zhang L, Liu Y, Cao X, Ye Y, Cheng MM, Zheng G (2018) Crowd counting
  with deep negative correlation learning. In: Proceedings of the IEEE
  Conference on Computer Vision and Pattern Recognition, pp 5382--5390

\bibitem[{Simonyan and Zisserman(2014)}]{simonyan2014very}
Simonyan K, Zisserman A (2014) Very deep convolutional networks for large-scale
  image recognition. arXiv preprint arXiv:14091556

\bibitem[{Sindagi and Patel(2017{\natexlab{a}})}]{sindagi2017cnn}
Sindagi VA, Patel VM (2017{\natexlab{a}}) Cnn-based cascaded multi-task
  learning of high-level prior and density estimation for crowd counting. In:
  Proceedings of the IEEE International Conference on Advanced Video and Signal
  Based Surveillance, pp 1--6

\bibitem[{Sindagi and Patel(2017{\natexlab{b}})}]{sindagi2017generating}
Sindagi VA, Patel VM (2017{\natexlab{b}}) Generating high-quality crowd density
  maps using contextual pyramid cnns. In: Proceedings of the IEEE International
  Conference on Computer Vision, pp 1879--1888

\bibitem[{Sindagi et~al.(2019)Sindagi, Yasarla, and Patel}]{sindagi2019pushing}
Sindagi VA, Yasarla R, Patel VM (2019) Pushing the frontiers of unconstrained
  crowd counting: New dataset and benchmark method. In: Proceedings of the IEEE
  International Conference on Computer Vision, pp 1221--1231

\bibitem[{Sindagi et~al.(2020)Sindagi, Yasarla, and
  Patel}]{sindagi2020jhu-crowd++}
Sindagi VA, Yasarla R, Patel VM (2020) Jhu-crowd++: Large-scale crowd counting
  dataset and a benchmark method. Technical Report

\bibitem[{Walach and Wolf(2016)}]{walach2016learning}
Walach E, Wolf L (2016) Learning to count with cnn boosting pp 660--676

\bibitem[{Wan et~al.(2019)Wan, Luo, Wu, Chan, and Liu}]{wan2019residual}
Wan J, Luo W, Wu B, Chan AB, Liu W (2019) Residual regression with semantic
  prior for crowd counting. In: Proceedings of the IEEE Conference on Computer
  Vision and Pattern Recognition, pp 4036--4045

\bibitem[{Wang et~al.(2015)Wang, Zhang, Yang, Liu, and Cao}]{wang2015deep}
Wang C, Zhang H, Yang L, Liu S, Cao X (2015) Deep people counting in extremely
  dense crowds. In: Proceedings of the 23rd ACM international conference on
  Multimedia, ACM, pp 1299--1302

\bibitem[{Wang et~al.(2018{\natexlab{a}})Wang, Chen, Nie, and
  Li}]{wang2018detecting}
Wang Q, Chen M, Nie F, Li X (2018{\natexlab{a}}) Detecting coherent groups in
  crowd scenes by multiview clustering. IEEE Transactions on Pattern Analysis
  and Machine Intelligence \doi{10.1109/TPAMI.2018.2875002}

\bibitem[{Wang et~al.(2018{\natexlab{b}})Wang, Wan, and Yuan}]{Qi2017Deep}
Wang Q, Wan J, Yuan Y (2018{\natexlab{b}}) Deep metric learning for crowdedness
  regression. IEEE Transactions on Circuits and Systems for Video Technology
  28(10):2633--2643

\bibitem[{Wang et~al.(2019)Wang, Gao, Lin, and Yuan}]{wang2019learning}
Wang Q, Gao J, Lin W, Yuan Y (2019) Learning from synthetic data for crowd
  counting in the wild. In: Proceedings of IEEE Conference on Computer Vision
  and Pattern Recognition (CVPR), pp 8198--8207

\bibitem[{Wang et~al.(2020)Wang, Gao, Lin, and Li}]{gao2020nwpu}
Wang Q, Gao J, Lin W, Li X (2020) Nwpu-crowd: A large-scale benchmark for crowd
  counting. arXiv preprint arXiv:200103360

\bibitem[{Wang et~al.(2004)Wang, Bovik, Sheikh, and Simoncelli}]{wang2004image}
Wang Z, Bovik AC, Sheikh HR, Simoncelli EP (2004) Image quality assessment:
  from error visibility to structural similarity. IEEE transactions on image
  processing 13(4):600--612

\bibitem[{Xiong et~al.(2017)Xiong, Shi, and Yeung}]{xiong2017spatiotemporal}
Xiong F, Shi X, Yeung DY (2017) Spatiotemporal modeling for crowd counting in
  videos. arXiv: Computer Vision and Pattern Recognition

\bibitem[{Yan et~al.(2019)Yan, Yuan, Zuo, Tan, Wang, Wen, and
  Ding}]{yan2019perspective}
Yan Z, Yuan Y, Zuo W, Tan X, Wang Y, Wen S, Ding E (2019) Perspective-guided
  convolution networks for crowd counting. In: Proceedings of the IEEE
  International Conference on Computer Vision, pp 952--961

\bibitem[{Yuan et~al.(2014)Yuan, Fang, and Wang}]{yuan2014online}
Yuan Y, Fang J, Wang Q (2014) Online anomaly detection in crowd scenes via
  structure analysis. IEEE transactions on cybernetics 45(3):548--561

\bibitem[{Zhang et~al.(2016{\natexlab{a}})Zhang, Kang, Li, Wang, Xie, and
  Yang}]{zhang2016data}
Zhang C, Kang K, Li H, Wang X, Xie R, Yang X (2016{\natexlab{a}}) Data-driven
  crowd understanding: a baseline for a large-scale crowd dataset. IEEE
  Transactions on Multimedia 18(6):1048--1061

\bibitem[{Zhang et~al.(2016{\natexlab{b}})Zhang, Zhou, Chen, Gao, and
  Ma}]{zhang2016single}
Zhang Y, Zhou D, Chen S, Gao S, Ma Y (2016{\natexlab{b}}) Single-image crowd
  counting via multi-column convolutional neural network. In: Proceedings of
  the IEEE conference on Computer Vision and Pattern Recognition, pp 589--597

\bibitem[{Zhao et~al.(2019)Zhao, Zhang, Zhang, and Zhang}]{Zhao_2019_CVPR}
Zhao M, Zhang J, Zhang C, Zhang W (2019) Leveraging heterogeneous auxiliary
  tasks to assist crowd counting. In: The IEEE Conference on Computer Vision
  and Pattern Recognition (CVPR), pp 12736--12745

\bibitem[{Zhu et~al.(2017)Zhu, Park, Isola, and Efros}]{zhu2017unpaired}
Zhu JY, Park T, Isola P, Efros AA (2017) Unpaired image-to-image translation
  using cycle-consistent adversarial networks. arXiv preprint

\bibitem[{Zuo et~al.(2018)Zuo, Wu, Lin, Zhang, and Yang}]{zuo2018learning}
Zuo W, Wu X, Lin L, Zhang L, Yang MH (2018) Learning support correlation
  filters for visual tracking. IEEE transactions on pattern analysis and
  machine intelligence 41(5):1158--1172

\end{thebibliography}

\end{document}